\documentclass{article} 
\usepackage{iclr2026_conference,times}

\usepackage{amsmath,amsfonts,bm}

\def\eqref#1{equation~\ref{#1}}

\def\1{\bm{1}}

\DeclareMathAlphabet{\mathsfit}{\encodingdefault}{\sfdefault}{m}{sl}
\SetMathAlphabet{\mathsfit}{bold}{\encodingdefault}{\sfdefault}{bx}{n}

\usepackage{arydshln}
\usepackage{graphicx}
\usepackage{booktabs} 
\usepackage{blindtext}
\usepackage{multicol}
\usepackage{url}
\usepackage{caption}
\usepackage{arydshln}
\usepackage{color}
\usepackage{xcolor}
\usepackage{wrapfig}
\usepackage{enumitem}
\usepackage{multirow} 
\usepackage[export]{adjustbox}
\usepackage{float}
\usepackage{epsfig}
\usepackage{amsmath}
\usepackage{amssymb} 
\usepackage{xparse} 
\usepackage{tabularx}
\usepackage{subcaption}
\usepackage{fancyhdr}
\usepackage[hang,flushmargin]{footmisc} 
\usepackage{pifont} 
\usepackage{subcaption}
\usepackage{boxedminipage}
\usepackage{framed}
\usepackage{soul}
\usepackage{xspace}
\usepackage{makecell}
\usepackage{subcaption}
\usepackage{vcell}
\usepackage{colortbl}
\usepackage[nodisplayskipstretch]{setspace}
\usepackage{seqsplit} 
\usepackage{array}
\usepackage{todonotes}
\usepackage{CJKutf8} 
\usepackage{verbatim}
\usepackage[most]{tcolorbox} 
\usepackage[skip=10pt plus 5pt minus 5pt]{parskip}
\usepackage{titlesec}
\usepackage[utf8]{inputenc} 
\usepackage[T1]{fontenc}    
\usepackage{algorithm}
\usepackage{algpseudocode}
\usepackage{amsfonts}       
\usepackage{nicefrac}       
\usepackage{microtype}      
\usepackage{xcolor}         
\usepackage{lscape}
\usepackage{tikz}
\usepackage{placeins}
\usepackage{tcolorbox}
\usepackage{listings}
\usepackage{pgfplots}
\usepackage{pgfplotstable}
\pgfplotsset{compat=1.17}
\usepackage{wasysym} 
\usepackage{verbatim}
\usepackage{fancyvrb}

\newcommand{\coloredblock}[1]{\textcolor{#1}{\rule{2mm}{2mm}}}

\definecolor{Math}{HTML}{F27970}
\definecolor{Physics}{HTML}{BB9727}
\definecolor{Chemistry}{HTML}{54B345}
\definecolor{Biology}{HTML}{32B897}
\definecolor{Geography}{HTML}{05B9E2}
\definecolor{Astronomy}{HTML}{8983BF}
\definecolor{Computer Science}{HTML}{C76DA2}
\definecolor{History}{HTML}{EFD496}
\definecolor{Law}{HTML}{DFB6BC}
\definecolor{easy}{HTML}{8ECFC9}
\definecolor{medium}{HTML}{FFBE7A}
\definecolor{hard}{HTML}{FA7F6F}

\definecolor{rule}{HTML}{eca680}
\definecolor{model}{HTML}{99B9E9}
\definecolor{answer}{HTML}{f9d580}
\definecolor{process}{HTML}{05B9E2}

\definecolor{lightblue}{RGB}{173, 216, 230}
\definecolor{darkblue}{HTML}{24AADD}

\definecolor{Single-modal to multi-model}{HTML}{DC8B70}
\definecolor{Manual annotation}{HTML}{7BBDB6}
\definecolor{Automatic annotation}{HTML}{A9A9A9}
\definecolor{mybrown}{RGB}{128,64,0}

\definecolor{shadecolor}{RGB}{237,237,237}
\definecolor{iris}{rgb}{0.35, 0.31, 0.81}
\definecolor{amaranth}{rgb}{0.9, 0.17, 0.31}

\definecolor{light-gray0}{gray}{0.92}
\definecolor{backred}{RGB}{250, 200, 200}
\definecolor{backblue}{RGB}{210, 230, 250}
\definecolor{backgreen}{RGB}{187,227,220}
\newcommand{\high}{\cellcolor{backblue}}
\newcommand{\light}{\cellcolor{backred}}
\newcommand{\midle}{\cellcolor{backgreen}}

\usepackage{hyperref}
\definecolor{Gold}{HTML}{D4AF37}
\definecolor{Silver}{HTML}{C0C0C0}
\definecolor{Bronze}{HTML}{CD7F32}

\title{{\huge \textcolor{blue!60!cyan}{\textbf{Q-Mirror}}}: Unlocking the Multi-Modal Potential of Scientific Text-Only QA Pairs}

\iclrfinalcopy 

\author{Junying Wang$^{1,2,*}$, Zicheng Zhang$^{2,*,\dagger}$, Ye Shen$^{2,3}$, Yalun Wu$^{3}$,  Yingji Liang$^{2}$,\\
\textbf{Yijin Guo$^{2,3}$, Farong Wen$^{2,3}$, Wenzhe Li$^{2}$, Xuezhi Zhao$^{2}$, Qi Jia$^{2}$, Guangtao Zhai$^{2,3,\dagger}$}
\\
{$^1$Fudan University, $^2$Shanghai Artificial Intelligence Laboratory, $^3$Shanghai Jiao Tong University}\\
{\footnotesize $^\dagger$Corresponding author.  \quad {AIBench Homepage: \url{https://aiben.ch/}}}\\
 {\footnotesize{Project Page: \url{https://github.com/aiben-ch/Q-Mirror}} }
}
\begin{document}

\maketitle

\begin{abstract}
High-quality, multi-modal benchmarks are crucial for advancing scientific reasoning in large models yet their manual creation is costly and unscalable. To address this bottleneck, we explore the potential for transforming Text-Only QA Pairs (TQAs) into high-quality Multi-Modal QA Pairs (MMQAs), which include three parts: 1) \textbf{Task Definition \& Evaluation Rubric}: We develop a {TQA-to-MMQA framework} and establish a {comprehensive, multi-dimensional {MMQA quality rubric}} that provides principles for the transformation. 2) \textbf{Benchmark Construction}: Then we construct two extensive benchmarks to rigorously evaluate state-of-the-art generation \& understanding models on the distinct tasks of {MMQA generation \& MMQA quality evaluation}. 3) \textbf{Preliminary Solution}: We develop an {agentic system} \textit{(Q-Mirror)}, which operationalizes our framework by integrating MMQA generation and evaluation into a closed loop for iterative refinement. Our experiments show that while state-of-the-art models can generate MMQAs, their outputs still leave substantial gaps, underscoring the need for reliable evaluation. We further demonstrate that top-tier understanding models align closely with human judgment in MMQA quality assessment. Leveraging both insights, the Q-Mirror agent raises average scores from 78.90 to 85.22 and pass rates from 72\% to 95\%, offering a practical path to large-scale scientific benchmarks.
\end{abstract}

\section{Introduction}
High-quality scientific data is the core to the benchmark of large scientific models~\citep{Intern-S1,SLLM,SciHorizon}. Since text-only data are relatively easy to collect and standardize, the community has built extensive banks of text-only QA pairs. However, real-world scientific problem solving is inherently multi-modal, often requiring the integration of visual diagrams, formulas, charts, and experimental setups. Thus, the demand for multi-modal scientific data is both clear and urgent. In response, several efforts have attempted to curate multi-modal benchmarks, yet the collection of such data is far more challenging, which requires higher annotation costs, richer domain expertise, and complex formatting. It has resulted in a pronounced imbalance: text-only resources greatly outnumber multi-modal ones. This observation motivates our guiding question: 

\quad \quad \quad \quad \quad \emph{Why not transform existing text-only QA pairs into multi-modal ones?}

Interestingly, many existing text-only resources already contain implicit multi-modal cues~\citep{T-SciQ,VLM2-Bench,MMCOT}. Geometry problems often reference diagram, physics questions describe experimental setups, and chemistry tasks mention molecular structures or reaction schemes~\citep{Geometry,Symbolic,Kuchemann2025,Yin_2024}. For instance, a physics problem asking to calculate the trajectory of a projectile, while described textually, implicitly calls for a diagram illustrating the initial velocity, angle, and gravitational force. These latent visual or structural elements, though expressed textually, could be surfaced and enriched to create multi-modal QA pairs (MMQAs). If such transformations were possible at scale, they would unlock the value of vast text-only banks without incurring the full cost of new multi-modal data collection. This perspective reframes the challenge: rather than treating text-only and multi-modal resources as disjoint, we explore how to systematically convert one into the other to accelerate multi-modal benchmark construction and, ultimately, advance scientific reasoning in large models.

Despite this potential, there is still no systematic framework for converting text-only scientific questions into multi-modal formats. Existing datasets treat text-only and multi-modal resources as largely disjoint, and current augmentation methods are limited to narrow tasks or small scales~\citep{aug_survey,data_aug}. Consequently, the vast repositories of TQAs remain underutilized, and the lack of scalable transformation tools constrains the development and evaluation of large scientific models. The rapid advancements in large language models (LLMs) and, more recently, large multi-modal models (LMMs), have opened unprecedented opportunities for complex content generation and understanding, making this systematic transformation more feasible than ever before.

Therefore, we explore the framework designed to transform TQAs into high-quality MMQAs. First, we establish a {principled quality rubric} to rigorously define what constitutes a successful transformation. Second, using this rubric as a foundation, we conduct two extensive benchmarks that evaluate state-of-the-art models on MMQA generation and evaluation, mapping out the current landscape of capabilities. Finally, armed with insights from these benchmarks, we develop a {novel agentic system} ({Q-Mirror}) that operationalizes our framework, orchestrating generation and judgment in a closed-loop process to autonomously produce high-quality, refined MMQAs. As illustrated in Figure~\ref{fig:outline}, these elements collectively motivate and shape our framework. Building on this foundation, our contributions can be summarized as follows:

\begin{figure}[t]
    \centering
    \vspace{-0.2cm}
    \includegraphics[width=0.95\linewidth]{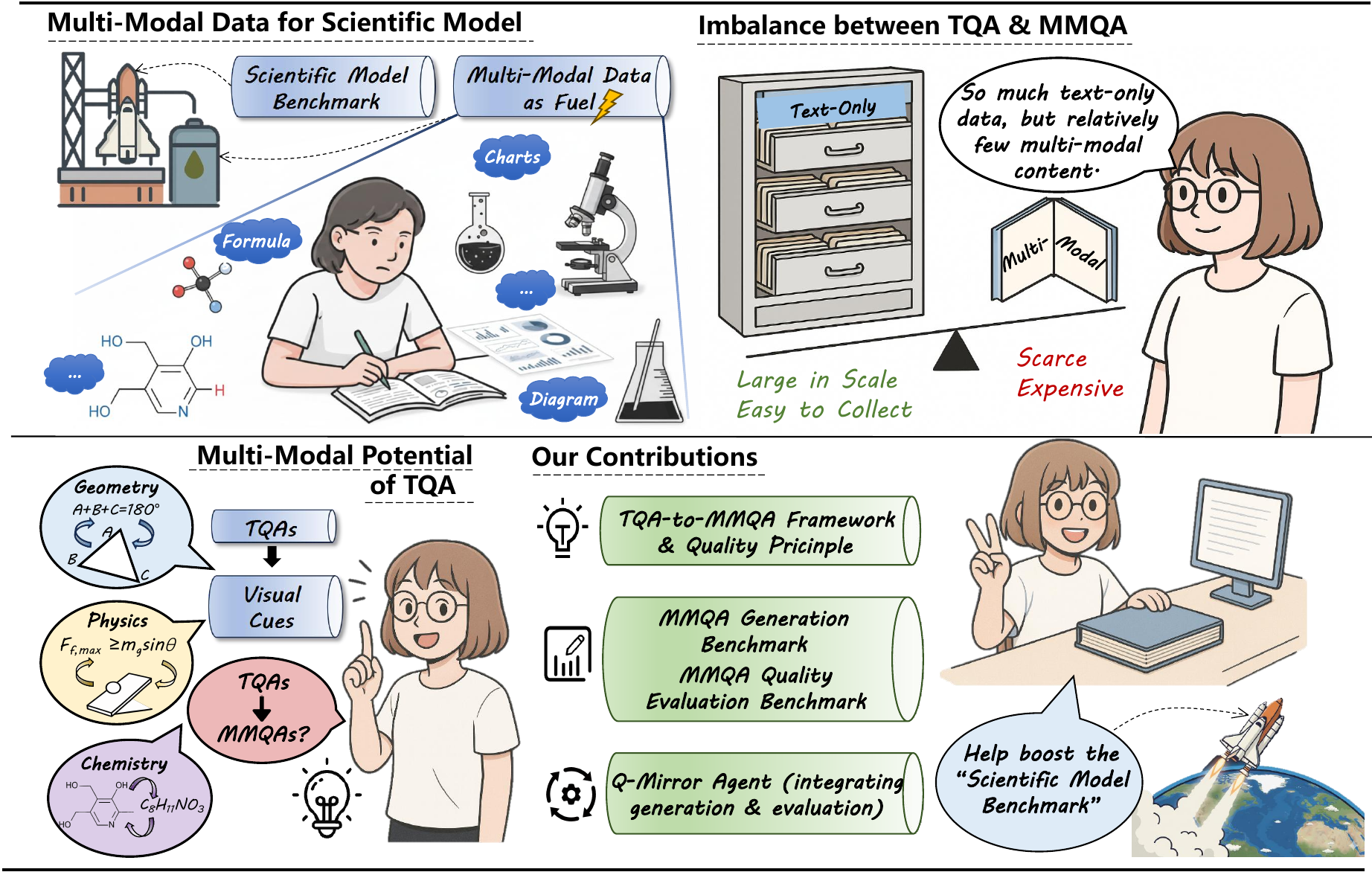}
    \vspace{-0.2cm}
    \caption{Overview of the motivation and key contributions, which illustrate: 1) the need for multi-modal data in scientific benchmark to advance model development, 2) the imbalance between text-only and multi-modal resources, 3) the latent potential of TQAs for multi-modal transformation, and 4) our contributions, including the transformation framework and quality principles, MMQA generation and evaluation benchmarks, and the Q-Mirror Agent for improving MMQA quality.}
    \label{fig:outline}
    \vspace{-0.2cm}
\end{figure}

\begin{itemize}[leftmargin=1em]
    \item \textbf{A Systematic TQA-to-MMQA Framework and Quality Rubric.} We propose the first systematic framework for the TQA-to-MMQA transformation, grounded in a comprehensive, multi-dimensional quality rubric. This rubric establishes principled criteria for high-quality MMQAs and serves as the foundation for entire study.

    \item \textbf{Benchmarks on MMQA Generation \& MMQA Quality Evaluation.} Grounded in our rubric, we construct and release two extensive benchmarks that evaluate state-of-the-art models on distinct tasks of 1) MMQA generation from TQAs, and 2) MMQA quality evaluation. These benchmarks reveal the capabilities of the current LLMs on complex generation and understanding.

    \item \textbf{An Agentic System for MMQA Generation \& Refinement.} Building upon the benchmark findings, we develop \textbf{Q-Mirror}, a novel agentic system that integrates MMQA generation and judgment. It operates in a closed-loop, iterative refinement process to autonomously convert TQAs into high-quality MMQAs at scale, providing a practical solution to the data scarcity problem.
\end{itemize}

\section{Related work}
\noindent\textbf{Scientific Benchmarks.} The advancement of scientific LLMs is fundamentally intertwined with the availability of high-quality data resources. 
Classical text-only resources such as TriviaQA~\citep{DBLP:journals/corr/JoshiCWZ17}, MMLU~\citep{hendrycks2020measuring}, C-Eval~\citep{huang2023c}, and GPQA~\citep{rein2024gpqa} cover diverse domains and reasoning levels at scale. 
In contrast, the recent emergence of LMMs has stimulated benchmarks like AI2D~\citep{kembhavi2016diagram}, ScienceQA~\citep{lu2022learn}, MathVista~\citep{lu2023mathvista}, and MMMU~\citep{yue2024mmmu}, which emphasize visual–language integration and cross-disciplinary coverage. 
However, as shown in Table~\ref{tab:bench-compare}, the number of instances in text-only benchmarks far exceeds those of multi-modal ones~\citep{multimodalbench,zhang2025lmmsurvey,Adv-CPG}. 
This imbalance highlights a structural gap: \emph{the rapid progress of scientific LMMs stands in tension with the limited scale and availability of multi-modal resources}.  

\noindent\textbf{Multi-Modal Data Construction.} Current multi-modal scientific resources are primarily \emph{manually curated}. 
For example, AI2D~\citep{kembhavi2016diagram} is constructed with expert-labeled diagrams and dense annotations, ScienceQA~\citep{lu2022learn} integrates web-harvested questions with human validation, and MathVista~\citep{lu2023mathvista} and MMMU~\citep{yue2024mmmu} merge domain-specific datasets through expert-driven curation. 
Although manual construction guarantees quality, it is inherently expensive, time-consuming, and difficult to scale~\cite{villalobos2022will,wang2025affordance}. 
While a few efforts explore automatic generation, such as ChemVLM~\citep{li2025chemvlm}, they remain limited in coverage and domain generality. 
These limitations motivate a new paradigm: \emph{systematically transforming abundant text-only resources into multi-modal ones}.  

\noindent\textbf{LLMs as Judges.} Parallel to benchmark construction, recent studies demonstrate the promise of LLMs as not only problem solvers but also \emph{judges} for evaluation~\citep{llmasajudge}. 
They can deliver rubric-grounded absolute scores with natural language explanations, or conduct pairwise preference ranking~\citep{wadhwa2025usingnaturallanguageexplanations, liu2025aligninghumanjudgementrole}. 
Their reliability is often stress-tested using counterfactual or perturbation-based methods, with consistency against human ratings quantified by correlation and calibration metrics~\citep{JudgeBench,ChatEval}. 
Building on this line of work, we introduce an \emph{agentic system} that integrates both generation and evaluation, allowing LMMs to iteratively refine multi-modal scientific QA pairs. 
This strategy extends the `LLM-as-Judge' paradigm and unlocks the latent potential of large-scale text-only corpora.

\begin{table}[!t]
\centering
\caption{Comparative Analysis of Scientific Benchmarks. This table contrasts key attributes of existing benchmarks to highlight \textbf{the scarcity of multi-modal resources and their high construction costs}, which motivates our work. \textbf{Legend:} \textit{Subjects}: \coloredblock{Math} Math, \coloredblock{Physics} Physics, \coloredblock{Chemistry} Chemistry, etc. 
\textit{Eval. Methods}: \coloredblock{rule} rule- / \coloredblock{model} model- / \coloredblock{answer} answer-level based. 
\textit{Difficulty}: \coloredblock{easy} Knowledge Recall, \coloredblock{medium} Concept Application, \coloredblock{hard} Cognitive Reasoning. 
\textit{\#Multi-Modal Data Construction}: \coloredblock{Manual annotation} Manual / \coloredblock{Automatic annotation} Automatic annotation.
}
\vspace{-0.2cm}
\renewcommand{\arraystretch}{1.05}
\resizebox{\linewidth}{!}{%
\begin{tabular}{llcllccccc}
\hline
Benchmark & Subjects & Multi-modal & Language & Size &\#Answer & Eval.  & Difficulty & \#Data. \\
\hline

C-Eval{\footnotesize~\citep{huang2023c}} & $\coloredblock{Math}$ $\coloredblock{Physics}$ $\coloredblock{Chemistry}$ $\coloredblock{Biology}$ $\coloredblock{Geography}$ $\coloredblock{Astronomy}$ $\coloredblock{Computer Science}$ $\coloredblock{History}$ & $\times$ & ZH & 13948 & 1 & $\coloredblock{rule}$ / $\coloredblock{answer}$  & \begin{tikzpicture}[scale=0.5] \fill[easy] (0,0) rectangle (0.51,0.20); \fill[medium] (0.51,0) rectangle (1.628,0.20); \fill[hard] (1.628,0) rectangle (2,0.20); \end{tikzpicture} & - \\

MATH{\footnotesize~\citep{MATH}} & $\coloredblock{Math}$ & $\times$ & EN  & 12500 & 1 & $\coloredblock{rule}$ / $\coloredblock{answer}$& \begin{tikzpicture}[scale=0.5] \fill[easy] (0,0) rectangle (0.52,0.20); \fill[medium] (0.52,0) rectangle (1.42,0.20); \fill[hard] (1.42,0) rectangle (2,0.20); \end{tikzpicture} & - \\

SciEval{\footnotesize~\citep{SciEval}} & $\coloredblock{Physics}$ $\coloredblock{Chemistry}$ $\coloredblock{Biology}$ & $\times$ & EN & 15901 & 4 & $\coloredblock{rule}$ / $\coloredblock{answer}$ &  \begin{tikzpicture}[scale=0.5] \fill[easy] (0,0) rectangle (0.71,0.20); \fill[medium] (0.71,0) rectangle (1.532,0.20); \fill[hard] (1.532,0) rectangle (2,0.20); \end{tikzpicture}  &- \\

SuperGPQA{\footnotesize~\citep{SuperGPQA}} & $\coloredblock{Math}$ $\coloredblock{Physics}$ $\coloredblock{Chemistry}$ $\coloredblock{Biology}$ $\coloredblock{Geography}$ $\coloredblock{Astronomy}$ $\coloredblock{Computer Science}$ $\coloredblock{History}$ $\coloredblock{Law}$ & $\times$ & EN & 26529 & 3 & $\coloredblock{rule}$ / $\coloredblock{answer}$ & \begin{tikzpicture}[scale=0.5] \fill[easy] (0,0) rectangle (0.5,0.20); \fill[medium] (0.5,0) rectangle (1.378,0.20); \fill[hard] (1.378,0) rectangle (2,0.20); \end{tikzpicture} & - \\

MMLU{\footnotesize~\citep{hendrycks2020measuring}} & $\coloredblock{Math}$ $\coloredblock{Physics}$ $\coloredblock{Chemistry}$ $\coloredblock{Biology}$ $\coloredblock{Geography}$ $\coloredblock{Astronomy}$ $\coloredblock{Computer Science}$ $\coloredblock{History}$ $\coloredblock{Law}$ & $\times$ & EN & 15908 & 1 & 
$\coloredblock{rule}$ / $\coloredblock{answer}$ & 
\begin{tikzpicture}[scale=0.5] \fill[easy] (0,0) rectangle (0.504,0.20); \fill[medium] (0.504,0) rectangle (1.718,0.20); \fill[hard] (1.718,0) rectangle (2,0.20); \end{tikzpicture} & - \\

CMMLU{\footnotesize~\citep{CMMLU}} & $\coloredblock{Math}$ $\coloredblock{Physics}$ $\coloredblock{Chemistry}$ $\coloredblock{Biology}$ $\coloredblock{Geography}$ $\coloredblock{Astronomy}$ $\coloredblock{Computer Science}$ $\coloredblock{History}$ $\coloredblock{Law}$ & $\times$ & ZH & 11528 & 1 & $\coloredblock{rule}$ / $\coloredblock{answer}$ & 
\begin{tikzpicture}[scale=0.5] \fill[easy] (0,0) rectangle (0.72,0.20); \fill[medium] (0.72,0) rectangle (1.8,0.20); \fill[hard] (1.8,0) rectangle (2,0.20); \end{tikzpicture} & - \\

MMLU-Pro{\footnotesize~\citep{MMLU-Pro}} & $\coloredblock{Math}$ $\coloredblock{Physics}$ $\coloredblock{Chemistry}$ $\coloredblock{Biology}$ $\coloredblock{Geography}$ $\coloredblock{Astronomy}$ $\coloredblock{Computer Science}$ $\coloredblock{History}$ $\coloredblock{Law}$ & $\times$ & EN & 12032 & 1 & $\coloredblock{rule}$ / $\coloredblock{answer}$ &  \begin{tikzpicture}[scale=0.5] \fill[easy] (0,0) rectangle (0.34,0.20); \fill[medium] (0.34,0) rectangle (1.655,0.20); \fill[hard] (1.655,0) rectangle (2,0.20); \end{tikzpicture} & - \\

TriviaQA{\footnotesize~\citep{DBLP:journals/corr/JoshiCWZ17}} & $\coloredblock{Physics}$ $\coloredblock{Chemistry}$ $\coloredblock{Biology}$ $\coloredblock{Geography}$ $\coloredblock{History}$ & $\times$ & EN &95956 &3 &\coloredblock{rule} / \coloredblock{answer} &- &-\\

EESE{\footnotesize~\citep{eese}} & $\coloredblock{Math}$ $\coloredblock{Physics}$ $\coloredblock{Chemistry}$ $\coloredblock{Biology}$ $\coloredblock{Geography}$ $\coloredblock{Astronomy}$ $\coloredblock{Computer Science}$ $\coloredblock{History}$ $\coloredblock{Law}$ & $\times$ &EN &100000+ & 5 & $\coloredblock{rule}$ / $\coloredblock{answer}$ & \begin{tikzpicture}[scale=0.5] \fill[easy] (0,0) rectangle (1.086,0.20); \fill[medium] (1.086,0) rectangle (1.774,0.20); \fill[hard] (1.774,0) rectangle (2,0.20); \end{tikzpicture} &-\\

\cdashline{1-9}
AI2D{\footnotesize~\citep{kembhavi2016diagram}} & $\coloredblock{Biology}$ $\coloredblock{Geography}$ $\coloredblock{Astronomy}$ & $\checkmark$ & EN & 15000+ & 1 & $\coloredblock{rule}$ / $\coloredblock{answer}$ / $\coloredblock{model}$  & - & $\coloredblock{Manual annotation}$  \\

ScienceQA{\footnotesize~\citep{lu2022learn}} & $\coloredblock{Physics}$ $\coloredblock{Chemistry}$ $\coloredblock{Biology}$ $\coloredblock{Geography}$ $\coloredblock{Astronomy}$ $\coloredblock{History}$& $\checkmark$ & EN & 21208 & 1 & $\coloredblock{rule}$ / $\coloredblock{answer}$  & \begin{tikzpicture}[scale=0.5] \fill[easy] (0,0) rectangle (1.31,0.20); \fill[medium] (1.31,0) rectangle (1.85,0.20); \fill[hard] (1.85,0) rectangle (2,0.20); \end{tikzpicture} & $\coloredblock{Manual annotation}$  \\

MathVista{\footnotesize~\citep{lu2023mathvista}} & $\coloredblock{Math}$ $\coloredblock{Physics}$ $\coloredblock{Biology}$& $\checkmark$ & EN & 6141 & 4 & $\coloredblock{rule}$ / $\coloredblock{answer}$ / $\coloredblock{model}$ & \begin{tikzpicture}[scale=0.5] \fill[easy] (0,0) rectangle (0.97,0.20); \fill[medium] (0.97,0) rectangle (1.586,0.20); \fill[hard] (1.586,0) rectangle (2,0.20); \end{tikzpicture} & $\coloredblock{Manual annotation}$ \\

MMMU{\footnotesize~\citep{yue2024mmmu}} & $\coloredblock{Math}$ $\coloredblock{Physics}$ $\coloredblock{Chemistry}$ $\coloredblock{Biology}$ $\coloredblock{Geography}$ $\coloredblock{Astronomy}$ $\coloredblock{Computer Science}$ $\coloredblock{History}$ & $\checkmark$ & EN & 11500 & 2 & $\coloredblock{rule}$ / $\coloredblock{answer}$  & \begin{tikzpicture}[scale=0.5] \fill[easy] (0,0) rectangle (0.282,0.20); \fill[medium] (0.282,0) rectangle (1.624,0.20); \fill[hard] (1.624,0) rectangle (2,0.20); \end{tikzpicture} &  $\coloredblock{Manual annotation}$  \\

ChemVLM{\footnotesize~\citep{li2025chemvlm}} & $\coloredblock{Chemistry}$ & $\checkmark$ &EN\&ZH & 4700+ &4 &\coloredblock{rule} / \coloredblock{answer} &- &\coloredblock{Automatic annotation}\\

HLE{\footnotesize~\citep{HLE}} & $\coloredblock{Math}$ $\coloredblock{Physics}$ $\coloredblock{Chemistry}$ $\coloredblock{Biology}$   $\coloredblock{Computer Science}$ & $\checkmark$ & EN & 3000 & 4 & $\coloredblock{answer}$ & \begin{tikzpicture}[scale=0.5] \fill[easy] (0,0) rectangle (0.4,0.20); \fill[medium] (0.4,0) rectangle (1.3,0.20); \fill[hard] (1.3,0) rectangle (2,0.20); \end{tikzpicture} & $\coloredblock{Manual annotation}$ \\
\hline
\end{tabular}}
\vspace{-0.3cm}
\label{tab:bench-compare}
\end{table}

\section{Task Definition \& Evaluation Rubric}
\label{sec:benchamrk}

To systematically realize TQAs into high-quality MMQAs, 
we first introduce a general framework for TQA-to-MMQA transformation (\S\ref{sec:framework}).
Next, we define a rubric that establishes principled criteria for high-quality MMQAs (\S\ref{sec:rubric}).

\begin{table}[t]\small
\centering
\caption{Brief description for the MMQA quality rubric, which includes three principles: Information Consistency (IC), Cross Modal Integration (CM), and Standalone Quality (QT).}
\label{tab:rubric_summary}
\vspace{-0.3cm}
\renewcommand{\arraystretch}{1.0}
\renewcommand\tabcolsep{10pt} 
\resizebox{\linewidth}{!}{\begin{tabular}{@{}p{0.08\linewidth} p{0.3\linewidth} p{0.1\linewidth} p{0.52\linewidth}@{}}
\toprule
\textbf{Principle} & \textbf{Metric} & \textbf{Notation} & \textbf{Description} \\
\midrule

\multirow{2}{*}{\textbf{IC}} 
& IL / Information Loss & $p_{-}$ & \textit{If critical information from the original TQA is lost.} \\
& IA / Information Addition & $p_{+}$ & \textit{If critical information from the original TQA is added.} \\
\midrule

\multirow{3}{*}{\textbf{CM}}
& SI / Solvability with Image & $p_s(I)$ & \textit{If the question can be solved from the image alone.} \\
& SQ / Solvability with Question & $p_s(T)$ & \textit{If the question can be solved from the text alone.} \\
& RE / Redundancy-Synergy & $f_s$ & Evaluates the degree of information overlap. \\
\midrule

\multirow{4}{*}{\textbf{QT}}
& NE / Natural Expression & $p_{\text{nat}}$ & \textit{If the text is linguistically fluent and coherent.} \\
& TQ / Technical Quality & $p_{\text{tech}}$ & \textit{If the image is technically correct and artifact-free.} \\
& AQ / Aesthetic Quality & $p_{\text{aes}}$ & \textit{If the image is visually clear and appealing.} \\
& SC / Semantical Clarity & $p_{\text{sem}}$ & \textit{If the image is plausible and scientifically sound.} \\
\bottomrule
\end{tabular}}
\vspace{-0.3cm}
\end{table}

\subsection{Framework for TQA-to-MMQA Transformation}
\label{sec:framework}
To unlock the latent multi-modal potential within existing text-based resources, we first establish a scalable and well-controlled process for systematically converting TQAs into MMQAs.


\textit{1) The Modal Conversion Stage.} For the given TQA, we use an LMM to first identify the parts that can be transformed into visual information. The LMM then removes these segments from the text and replaces them with an image placeholder, ensuring that the resulting question becomes visually dependent. Meanwhile, the LMM generates a detailed visual description based on the removed content, which serves as input for the text-to-image (T2I) model (\textit{detailed prompts in Appendix~\ref{sec:prompt_generation}}).

\textit{2) The Image Generation Stage.} A T2I model is utilized to synthesize the image using the detailed description from the previous stage. To ensure semantic alignment and stylistic coherence, we emphasize that the description generator (LMM) and the image generator (T2I model) should ideally belong to the \textbf{same model family} (e.g., \textit{GPT-4.1-2025-04-14} \citep{openai_gpt41_2025} and \textit{GPT-Image-1} \citep{openai_imagegen_2025}). Such family-level consistency often facilitates a closer coupling between textual semantics and visual synthesis, helping to reduce modal gaps and improve the generation quality.

\subsection{Quality Principles for MMQA Transformation}
\label{sec:rubric}

A good MMQA transformation should satisfy three key criteria. First, it must maintain \textit{information consistency}, ensuring that the essential meaning of the original TQA is preserved without losing or introducing critical details. Second, it should achieve \textit{cross-modal integration}, where the question and image provide complementary information such that neither modal alone is sufficient for solving the problem, and true multi-modal dependence is enforced. Third, it must guarantee high \textit{standalone quality}, with the text remaining fluent and natural, and the image being technically correct, visually clear, and scientifically sound.
Therefore, we establish a multi-dimensional rubric grounded in three principles: information consistency, cross-modal integration, and standalone quality (\textit{Table~\ref{tab:rubric_summary}}).

 \subsubsection{Preliminaries}
Let $Q_s$ be an original TQA and $Q_m = (T, I)$ be its transformed counterpart. The evaluation relies on a set of boolean predicates $\mathbb{I}[\cdot]$, which return true (1) or false (0). The composite scores for each principle are defined below, scaled from 0 to 100. To ensure reproducibility, each predicate is evaluated against specific operational criteria (\textit{details can be referred to in Appendix~\ref{sec:human_eval_protocol}}).

\begin{figure}[t]
    \centering
    \vspace{-0.2cm}
    \includegraphics[width=\linewidth]{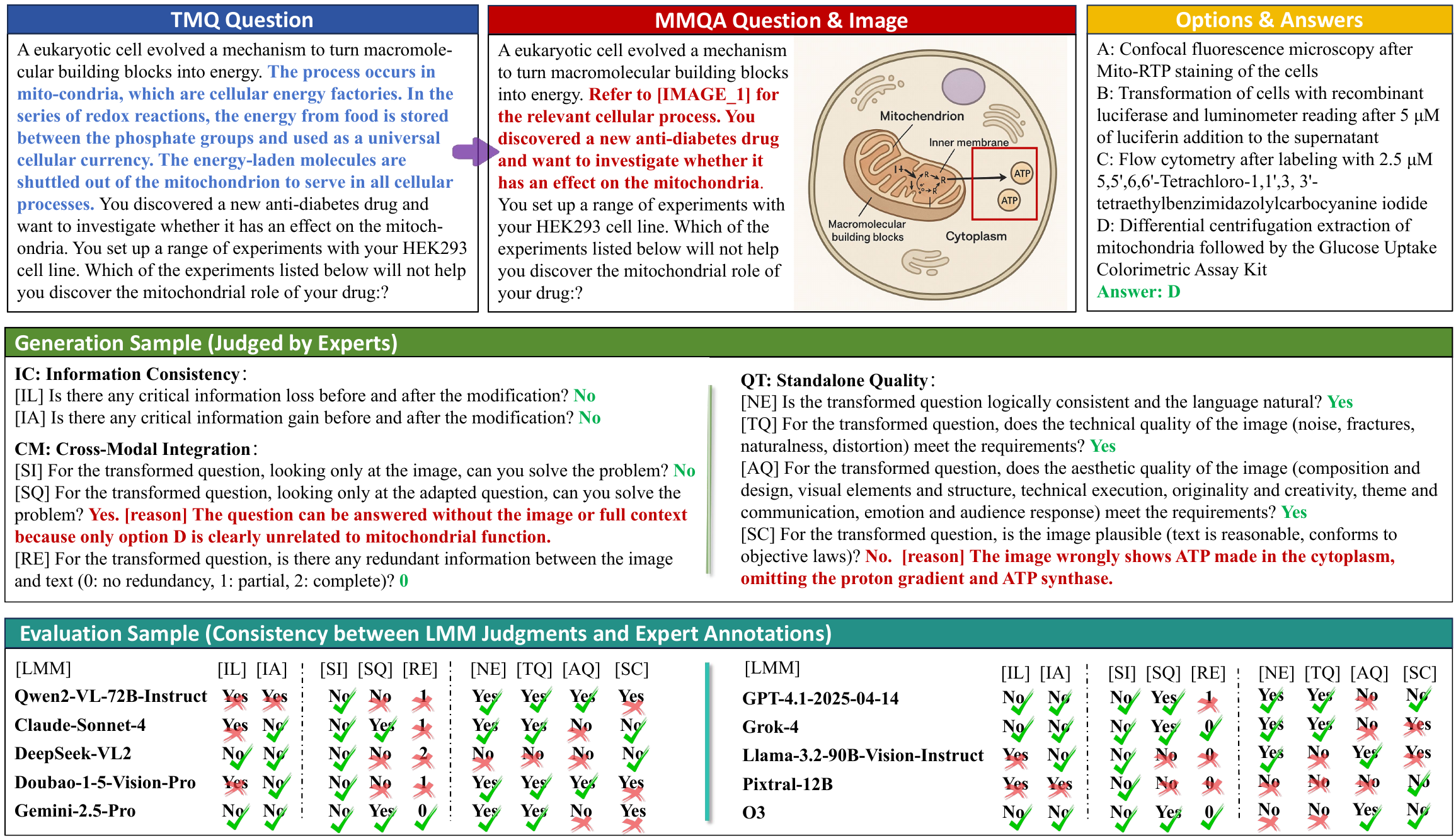}
    \vspace{-0.2cm}
    \caption{An illustration of the TQA-to-MMQA transformation, expert annotation, and LMM judge evaluation, including: 1) the full conversion of a TQA into an MMQA, 2) the expert annotation sample for the corresponding MMQA generation case, based on the proposed quality rubric, and 3) the evaluation results of LMMs with their correctness indicated against expert annotations.}
    \label{fig:case_main}
     \vspace{-0.35cm}
\end{figure}

\subsubsection{Evaluation Principles}
\label{sec:principle}

\textit{Principle 1: Information Consistency (IC).}
This principle measures semantic fidelity between the source TQA $Q_s$ and its transformed MMQA $Q_m$. 
We define a predicate set $\mathcal{P}_{\text{IC}} = \{p_{-}, p_{+}\}$, 
where $p_{-}$ detects missing critical information and $p_{+}$ detects spurious additions. 
The IC score is the complement of their violation rate:
\begin{equation}
    \text{IC}(Q_m, Q_s) =  1 - \frac{1}{|\mathcal{P}_{\text{IC}}|} 
    \sum_{p \in \mathcal{P}_{\text{IC}}} \mathbb{I}[p(Q_m, Q_s)],
\end{equation}

\textit{Principle 2: Cross-Modal Integration (CM).}
This principle evaluates whether text and image contribute complementary, indispensable information. 
We test single-modal solvability using $p_s(T)$ and $p_s(I)$, 
and assess synergy via $f_s(T, I) \in \{\textit{Complete, Partial, None}\}$. 
Let $\beta$ denote the weight mapping of overlap categories. 
The CM score is defined as:
\begin{equation}
    \text{CM}(Q_m) = \frac{1}{3} \Big( (1 - \mathbb{I}[p_s(T)]) 
    + (1 - \mathbb{I}[p_s(I)]) 
    + \beta[f_s(T, I)] \Big),
\end{equation}
where we set $\beta[\textit{Partial}]=0.75$, $\beta[\textit{None}]=0.25$, and $\beta[\textit{Complete}]=0$.

\textit{Principle 3: Standalone Quality (QT).}
This principle assesses the intrinsic quality of each modal. 
We define $\mathcal{P}_{\text{QT}} = \{p_{\text{nat}}(T), p_{\text{tech}}(I), p_{\text{aes}}(I), p_{\text{sem}}(I)\}$, 
covering textual fluency and visual adequacy. 
The QT score is the average success rate:
\begin{equation}
    \text{QT}(Q_m) = \frac{1}{|\mathcal{P}_{\text{QT}}|} 
    \sum_{p \in \mathcal{P}_{\text{QT}}} \mathbb{I}[p(Q_m)].
\end{equation}

\textit{Final Aggregated Score (AVG).}
The overall score is a weighted sum of the three principles:
\begin{equation}
    \text{AVG}= \alpha_{\text{IC}} \cdot \text{IC}(Q_m, Q_s) 
    + \alpha_{\text{CM}} \cdot \text{CM}(Q_m) 
    + \alpha_{\text{QT}} \cdot \text{QT}(Q_m),
\end{equation}
with $\alpha_{\text{IC}} + \alpha_{\text{CM}} + \alpha_{\text{QT}} = 1$. 
We set $\alpha_{\text{IC}}=0.3$, $\alpha_{\text{CM}}=0.3$, and $\alpha_{\text{QT}}=0.4$, 
reflecting that standalone quality is a prerequisite for usability. 

\section{Benchmark Construction}
In this section, we first introduce the TQA collection (\S\ref{sec:tqa_coll}). Then, we construct a benchmark for TQA-to-MMQA generation (\S\ref{sec:benchmark_genera}),
and finally, we validate the capabilities of popular LMMs as automated judges for MMQA quality evaluation (\S\ref{sec:judge_validation}).
A corresponding case is shown in {Figure~\ref{fig:case_main}}, which illustrates the TQA-to-MMQA transformation, expert annotation based on the proposed rubric, and the comparison of LMM judgments with human annotations. 

\subsection{TQA Collection}
 \label{sec:tqa_coll}
We begin by sampling candidate TQAs from multiple authoritative scientific benchmarks \citep{rein2024gpqa,eese} to ensure diversity and disciplinary breadth. An LMM is then employed to automatically detect which questions are suitable for transformation into multi-modal form, discarding those that lack clear visualizable elements. Next, human experts review the remaining questions to verify their scientific value, refine their formulations, and guarantee balanced coverage across 22 scientific disciplines. Finally, the curated pool is partitioned into two subsets according to difficulty, determined jointly by dataset-provided difficulty labels and expert judgment: \textbf{Q-Mirror-Expert (310 questions)} focuses on frontier or interdisciplinary concepts that often lack standard visual representations, while \textbf{Q-Mirror-Grad (130 questions)} targets well-established graduate-level knowledge requiring precise visual reproduction of structured scientific information. Further details on TQA Collection are provided in Appendix~\ref{sec:dataset_details}.

\subsection{MMQA Generation Benchmark}
\label{sec:benchmark_genera}

To establish a gold-standard benchmark for MMQA generation, we institute a rigorous human evaluation protocol grounded in our core principles (§\ref{sec:rubric}). We assemble a panel of 50 domain experts, primarily researchers in STEM fields, who experience an intensive training and calibration process. Each model-generated MMQA is independently assessed by at least two experts using the fine-grained rubric that is based on our principles. Experts assign multi-dimensional scores and provide textual justifications for any identified flaws. This effort culminates in the MMQA generation benchmark, a dataset enriched with expert-verified quality labels. A complete formalization of our rubric, including detailed guidelines and examples provided to annotators, is available in Appendix~\ref{sec:prompt_generation}.

\subsection{MMQA Evaluation Benchmark}
\label{sec:judge_validation}
While human annotation provides the gold standard, its high cost and low scalability are critical bottlenecks for iterative development. To overcome this, we systematically evaluate the viability of state-of-the-art LMMs as scalable, automated judges. We prompt a diverse suite of ten LMMs to score the MMQAs from our benchmark. Each LMM judge is provided with a structured prompt containing the detailed multi-dimensional rubric (identical to the one used by human experts), and a required JSON output format. We then measure the alignment between the LMM-generated scores and the human ground truth. The complete prompt templates are provided in Appendix~\ref{sec:prompt_eval}.

\begin{algorithm}[!t]\small
\caption{Q-Mirror: Iterative Refinement Workflow}
\label{alg:q_mirror}
\begin{algorithmic}[1]
\State \textbf{Input:} Original TQA $\mathcal{Q}_s$, Quality Threshold $\tau$, Max Iterations $N$, Candidates per Iteration $K$.
\State \textbf{Output:} High-quality MMQA $\mathcal{Q}_m^*$.

\State Initialize $feedback \gets \text{null}$, $\text{best\_score} \gets -\infty$, $\mathcal{Q}_m^* \gets \text{null}$
\For{$i \gets 1$ to $N$}
    \State $\text{Candidates} \gets \textbf{Planner.GenerateCandidates}(\mathcal{Q}_s, feedback, K)$
    
    \State $\text{best\_cand}, \text{best\_score\_curr}, \text{judgments} \gets \textbf{Evaluation.EvaluateBatch}(\text{Candidates})$
    
    \If{$\text{best\_score\_curr} > \text{best\_score}$}
        \State $\text{best\_score} \gets \text{best\_score\_curr}$
        \State $\mathcal{Q}_m^* \gets \text{best\_cand}$
    \EndIf
    
    \If{$\text{best\_score} \ge \tau$} \textbf{break} \EndIf
    
    \State $feedback \gets \textbf{Controller.GenerateFeedback}(\text{judgments})$
\EndFor
\State \textbf{return} $\mathcal{Q}_m^*$
\end{algorithmic}
\end{algorithm}

\section{Preliminary Solution}
\label{sec:methodology}
Building upon the validated reliability of automated judges, we present the complete Q-Mirror Agent. It is an autonomous system that orchestrates generation and refinement within a closed-loop workflow, as detailed in Algorithm~\ref{alg:q_mirror}. The details are as follows:

\textit{1) The Planner Stage.} The Q-Mirror begins with the {Planner}, a module that produces $K$ candidate MMQAs from a given TQA. 
For each candidate, the Planner simultaneously 1) reformulates the textual component through a process that involves identifying parts best converted to visuals, replacing them with explicit placeholders, and rephrasing the context, and 2) generates a detailed visual description. This description is then immediately converted into a rendered image via a coupled T2I model, 
yielding a complete MMQA. Importantly, the Planner is feedback-aware: in subsequent iterations, it conditions its generation not only on the source TQA 
but also on structured revision signals provided by the Controller, enabling progressive improvement. 

\textit{2) The Evaluator Stage.} The resulting candidates are then subjected to the {Evaluator}, carried out by $M$ top-performing LMMs ranked in our benchmark (\S\ref{sec:judge_validation}). 
The ensemble not only assigns rubric-based quality scores but also provides qualitative judgments on critical flaws such as semantic omissions, factual inaccuracies, or weak cross-modal alignment, together with prescriptive suggestions on how these issues can be refined in the next iteration. 
If the best candidate already exceeds the predefined quality threshold $\tau$, it is accepted as the final output. 
Otherwise, the collective feedback of the evaluators serves as the basis for refinement.

\textit{3) The Controller Stage.} Finally, the Controller synthesizes the feedback of the ensemble into actionable, structured revision instructions. 
Rather than aggregating scores mechanically, the Controller prioritizes high-severity issues (e.g., factual errors) over superficial aspects (e.g., stylistic variation), 
thereby guiding the Planner toward substantive improvements. 
The refined instructions are then returned to the Planner, which regenerates a new batch of candidates. 
This cycle of \emph{generation–evaluation–revision} continues until a candidate surpasses the quality threshold or the maximum number of iterations $N$ is reached.

\begin{table*}[t]
\centering
 \vspace{-0.2cm}
\caption{Performance of model families (\textit{the description generator and image generator from the same family, details in \S\ref{sec:exp_setup})}) on MMQA generation benchmark, top performance highlighted. The scores are all rescaled to a 0-100 range for presentation. }
\label{tab:generation}
 \small
 \resizebox{1.0\linewidth}{!}{
    \begin{tabular}{l|ccc|cccc|ccccc|cc}
    \toprule
    \textbf{Model Family}   & \textit{IL} & \textit{IA} & \textit{\textbf{IC}} & \textit{SI} & \textit{SQ} & \textit{RE} & \textit{\textbf{CM}} & \textit{NE} & \textit{TQ} & \textit{AQ} & \textit{SC} & \textit{\textbf{QT}}  & \textbf{AVG} & \textbf{RK}\\ 
    \midrule
    \multicolumn{15}{l}{ \textit{Performance on Q-Mirror-Expert}} \\ \hline

    Doubao Family  & 61.54 &\high{85.39}	&73.46 &\high{100.00}	&\high{90.65} &18.55	&69.73	&\high{99.68}	&77.69	&69.23	& 29.23	&68.96	&70.54 &4\\
    Qwen2 Family   & \high{76.15}  &73.08	&74.62	&99.03	&83.23	&51.92	&78.06 &97.74	&72.46	&71.46	&\high{33.08}	&68.69	&73.28 & 2 \\
    Grok Family  & 62.31 &75.48 &68.90	&99.03 &85.48	&60.40	&81.64 &90.97	&75.00	&69.23	&26.92	&65.53	&71.37 &3\\
    GPT Family & 71.54  &80.77	&\high{76.15} &\high{100.00}	&88.71 &\high{71.94}	&\high{86.88}	&98.07	&\high{80.00}	&69.23	&26.92	&\high{68.55}	&\high{76.33} & 1  \\
    Qwen-VL Family &64.96 &71.47 &68.22	&\high{100.00}	&80.72	&51.95	&77.56	&98.70	&75.35	&59.69	&27.09	&65.21	&69.81 &5\\
    \rowcolor{light-gray0} Q-Mirror Agent &79.29 &84.36 &81.83	&100 &91.58	&75.49	&89.03	&100	&82.63	&78.69	&52.65	&78.49	&82.65 &/ \\
      \midrule
    \multicolumn{15}{l}{ \textit{Performance on Q-Mirror-Grad} } \\ \hline

    Doubao Family  & 88.39 &87.42	&87.90 &\midle{100.00}	&90.77 &45.19	&78.65	&\midle{100.00}	&79.36	&79.03	& 30.00	&68.96	&77.55 & 2 \\
    Qwen2 Family   & 85.48  &73.87	&79.68	&99.23	&90.77	&55.24	&81.75	&\midle{100.00}	&78.71	&78.39	&\midle{39.03}	&68.69	&75.90 &3\\
    Grok Family  & 80.65 &80.00 &80.32	&\midle{100.00} &\midle{93.08}	&60.85	&84.64 &97.69	&81.61	&75.16	&32.26	&65.53	&75.70 &4\\
    GPT Family & 83.23  &\midle{96.13}	&\midle{89.68} &\midle{100.00}	&89.23 &\midle{72.69}	&\midle{87.31}	&\midle{100.00}	&77.74	&\midle{80.32}	&29.44	&68.55	&\midle{80.52} & 1  \\
    Qwen-VL Family &85.44 &72.25 &78.84	&97.67	&91.47	&60.41	&83.19	&99.23	&81.43	&72.96	&28.66	&65.21	&74.69 &5\\

   \rowcolor{light-gray0} Q-Mirror Agent &87.98	&94.63	&91.30	&100	&92.15	&82.65	&91.60	&100	&84.97	&86.35	&57.67	&78.49	&86.267 &/ \\
      \midrule
    \multicolumn{15}{l}{ \textit{Performance on Q-Mirror-(Expert+Grad)} } \\ \hline

    Doubao Family  & 69.47 &\light{85.99}	&77.73 &\light{100.00}	&\light{90.68} &26.42	&72.37	&\light{99.77}	&78.18	&72.13	& 29.46	&72.10	&73.87 & 4 \\
    Qwen2 Family   & \light{78.91}  &73.31	&76.11	&99.09	&85.45	&52.90	&79.15 &98.41	&74.31	&73.51	&\light{34.84}	& \light{74.03}	&76.19  & 2 \\
    Grok Family  & 67.73 &76.82 &72.27	&99.32 &87.73	&60.53	&82.53 &92.95	&76.95	&70.98	&28.50	&71.68	&75.11 &3\\
    GPT Family & 74.99  &85.31	&\light{80.15} &\light{100.00}	&88.86 &\light{72.16}	&\light{87.01}	&98.64	&\light{79.33}	&\light{72.51}	&27.67	&71.88	&\light{78.90} & 1 \\
    Qwen-VL Family &71.01 &71.70 &71.36	&99.31	&83.89	&54.45	&79.22	&98.85	&77.15	&63.61	&27.55	&70.57	&73.40 &5\\
    \rowcolor{light-gray0}    Q-Mirror Agent &81.86	&87.40	&84.63	&100	&91.75	&77.61	&89.79	&100	&83.32	&80.95	&54.13	&82.25	&85.22 &/ \\
      \midrule
    \end{tabular}
    }
    \vspace{-0.3cm}
\end{table*}

\section{Experiments}

A series of experiments are conducted to validate the proposed framework, designed to answer three key questions: 1) How do the generation models perform on the TQA-to-MMQA transformation task? 2) How reliable are leading understanding models as MMQA quality judges compared to human experts? 3) How effectively does the proposed Q-Mirror agent improve MMQA quality?

\subsection{Experimental Setup}
\label{sec:exp_setup}

\paragraph{Models Families for MMQA Generation.}
For MMQA generation, we benchmark five diverse model families, including proprietary and prominent open-source leaders, including the \textbf{GPT Family} (\textit{GPT-4.1-2025-04-14}~\citep{openai_gpt41_2025}, \textit{GPT-Image-1}~\citep{openai_imagegen_2025}), the \textbf{Doubao Family} (\textit{Doubao-1-5-Thinking-Pro-250415}~\citep{bytedance_doubao_2024}, \textit{Doubao-Seedream-3-0-T2I-25041}~\citep{gao2025seedream}), the \textbf{Qwen2 Family} (\textit{Qwen2-VL-72B-Instruct}~\citep{wang2024qwen2}, \textit{Wanx2.1-T2I-Plus}~\citep{wan2025wan}), the \textbf{Grok Family} (\textit{Grok-4}~\citep{xai_grok4_2025}, \textit{Grok-2-Image}~\citep{xai_api_image_2025}), and \textbf{Qwen-VL Family}  (\textit{Qwen-VL-Max-Latest}~\citep{bai2023qwen}, \textit{Qwen-Image}~\citep{wu2025qwen}). By adopting intra-family model pairing to reduce architectural inconsistencies and modal gaps, this strategy enables a more accurate comparison of different model families in scientific T2I generation, particularly in evaluating semantic alignment and content faithfulness.

\paragraph{Models for MMQA Evaluation.}
For the MMQA quality judge evaluation, we assess ten powerful LMMs including \textit{Qwen2-VL-72B-Instruct}~\citep{wang2024qwen2}, \textit{Claude-Sonnet-4}~\citep{anthropic_claude4_2025}, \textit{DeepSeek-VL2}~\citep{deepseekvl2}, \textit{Doubao-1-5-Vision-Pro-250328}~\citep{bytedance_doubao_2024}, \textit{Gemini-2.5-Pro}~\citep{team2023gemini}, \textit{GPT-4.1-2025-04-14}~\citep{openai_gpt41_2025}, \textit{Grok-4}~\citep{xai_grok4_2025}, \textit{Llama-3.2-90B-Vision}~\citep{grattafiori2024llama}, \textit{Pixtral-12B}~\citep{agrawal2024pixtral}, and \textit{O3}~\citep{openai_o3_o4mini_2025}, to cover a range of architectures. All models are evaluated against our multi-dimensional rubric (\S\ref{sec:methodology}).

\paragraph{Q-Agent Parameters Setup.}
\label{sec:setup}
In practice, we set $\tau = 80.0$ (the acceptance threshold score), $M = 3$ (the number of top-performing LMMs in ensemble), $N = 5$ (the maximum number of refinement iterations), and $K = 4$ (the number of candidates generated per iteration) for Q-Mirror, balancing efficiency and reliability. In addition to reporting the AVG, the \textit{pass rate} is also defined as the proportion of MMQAs whose AVG exceeds the acceptance threshold $\tau = 80.0$. This metric reflects not only average improvements but also the robustness of the system in consistently producing high-quality outputs.
Moreover, the Q-Mirror agent \textit{Planner} is instantiated with \textit{GPT-Image-1}). The Evaluator module is an ensemble of the top-three performing judges identified in the evaluation benchmark: \textit{Grok-4}, \textit{GPT-4.1-2025-04-14}, and \textit{O3}. The specific prompts used for generation and evaluation are detailed in Appendix~\ref{sec:human_eval_protocol}, Appendix~\ref{sec:prompt_generation}, and Appendix~\ref{sec:prompt_eval}.

\begin{table*}[t!]
\centering
\caption{Performance on MMQA evaluation benchmark. 10 LMMs are evaluated by measuring the alignment of their scores with human expert annotations, with the best performance highlighted. The scores are all rescaled to a 0-100 range for presentation.}
\label{tab:judge}
\vspace{-0.2cm}
 \small
 \resizebox{1.0\linewidth}{!}
 {
    \begin{tabular}{l|ccc|cccc|ccccc|cc}
    \toprule
    \textbf{Model}   & \textit{IL} & \textit{IA} & \textit{\textbf{IC}} & \textit{SI} & \textit{SQ} & \textit{RE} & \textit{\textbf{CM}} & \textit{NE} & \textit{TQ} & \textit{AQ} & \textit{SC} & \textit{\textbf{QT}}  & \textbf{AVG} & \textbf{RK}\\ 
    \midrule
    \multicolumn{15}{l}{\textit{Judge Alignment on Q-Mirror-Expert}} \\ \hline
    Qwen2-VL-72B-Instruct  & 62.31 &31.94	&47.12 &95.81	&47.42 &26.13	&56.45	&39.03	&82.58	&31.29	& 48.39	&50.32	&51.20 & 8 \\
    Claude-Sonnet-4 &71.93 &46.92 &59.42	&96.77	&30.32	&38.06	&55.05	&53.42	&79.19	&47.90	&46.45	&56.74	&57.04 &5  \\
    DeepSeek-VL2  & 60.77 &31.54 &46.15	&91.85 &20.32	&24.19	&45.45 &36.45	&89.81	&37.42	&55.81	&43.23	&44.77 & 10 \\
    Doubao-1.5-Vision-Pro & 65.38  &37.15	&51.27 &97.92	&\high{61.61} &26.77	&\high{62.10}	&32.26	&59.29	&39.35	&57.10	&47.00	&52.81 &6\\
    Gemini-2.5-Pro &\high{72.69} &42.46 &57.58	&\high{99.68}	&50.97	&30.32	&60.32	&52.06	&87.61	&38.28	&51.61	&57.39	&58.33 &4\\
    GPT-4.1 & 70.00  &39.35	&54.68	&96.77	&45.48	&34.19	&58.82	&46.45	&\high{96.45}	&\high{57.10}	&49.32	& 62.33	&58.98 & 2 \\
    Grok-4 &71.54 &47.74 &59.64	&95.48	&26.77	&\high{42.90}	&55.05	&\high{54.62}	&83.87	&55.48	&\high{68.71}	&\high{65.67}	&\high{60.68} &1\\
    Llama-3.2-90B-Vision &63.08 &41.55 &52.31	&89.92	&29.35	&27.74	&49.01	&48.06	&87.71	&40.32	&38.39	&53.62	&51.84 &7\\
    Pixtral-12B &60.50 &39.26 &49.88	&90.58	&34.52	&20.97	&48.69	&44.85	&83.87	&26.77	&40.65	&49.03	&49.18 &9\\
    O3 &72.31 &\high{52.03} &\high{62.17}	&98.54	&42.10	&24.74	&55.13	&47.85	&82.90	&45.16	&57.74	&58.41	&58.55 &3 \\
      \midrule
    \multicolumn{15}{l}{\textit{Judge Alignment on Q-Mirror-Expert}} \\ \hline
    Qwen2-VL-72B-Instruct   & 74.10 &43.85	&58.97 &\midle{100.00}	&70.77 &33.08	&67.95	&42.77	&88.46	&35.71	& 49.31	&54.06	&59.70 & 7 \\
    Claude-Sonnet-4  &84.68 &51.94 &68.31	&\midle{100.00}	&\midle{73.08}	&40.77	&\midle{71.28}	&51.77	&80.77	&49.23	&49.54	&57.83	&65.01 &3 \\
    DeepSeek-VL2  & 77.39 &36.13 &56.76	&92.52 &67.25	&36.15	&65.31 &46.29	&89.38	&41.54	&56.54	&48.12	&55.87& 10 \\
    Doubao-1.5-Vision-Pro & 76.55  &42.77	&59.66 &\midle{100.00}	&70.77 &26.15	&65.64	&45.38	&63.08	&44.65	&56.92	&55.13	&59.64 &8\\
    Gemini-2.5-Pro &83.55 &47.85 &65.70	&99.68	&63.85	&26.92	&63.48	&52.23	&88.54	&37.69	&56.92	&58.85	&62.29 &5\\
    GPT-4.1  & 84.84  &\midle{63.85}	&\midle{74.34}	&\midle{100.00}	&71.54	&37.69	&69.74	&57.54	&\midle{97.69}	&\midle{62.31}	&60.15	& \midle{69.42}	&\midle{71.00} & 1  \\
    Grok-4 &82.71 &52.92 &67.82	&96.15	&65.38	&37.77	&51.58	&57.26	&91.77	&56.15	&\midle{70.77}	&68.99	&63.41 &4\\
    Llama-3.2-90B-Vision &73.52 &42.31 &57.91	&91.87	&56.11	&\midle{46.92}	&64.97	&49.23	&88.69	&56.92	&40.69	&58.88	&60.42 &6\\
    Pixtral-12B &74.52 &46.45 &60.48	&91.17	&68.16	&26.92	&62.08	&46.68	&86.89	&27.65	&51.62	&53.21	&58.05 &9\\
    O3 &\midle{86.87} &55.38 &71.13	&98.46	&68.77	&37.10	&68.11	&\midle{60.97}	&86.15	&52.73	&59.23	&64.77	&67.68 &2\\
      \midrule
     \multicolumn{15}{l}{\textit{Judge Alignment on Q-Mirror-(Expert+Grad)}} \\ \hline
    Qwen2-VL-72B-Instruct   & 65.79 &35.45	&50.62 &97.05	&54.32 &28.18	&59.85	&40.14	&84.32	&32.60	& 48.66	&51.43	&53.71 & 8 \\
    Claude-Sonnet-4   &75.69 &48.40 &62.05	&97.73	&42.95	&38.86	&59.85	&52.93	&79.66	&48.30	&47.36	&57.06	&59.39 &5 \\
    DeepSeek-VL2 & 65.68 &32.89 &49.29	&92.04 &34.19	&27.73	&51.32 &39.36	&89.68	&38.64	&56.02	&55.92	&52.55 &9 \\
    Doubao-1.5-Vision-Pro & 68.68  &38.81	&53.75 &98.54	&\light{64.32} &26.59	&63.15	&36.14	&60.41	&40.92	&57.05	&48.63	&54.52 &6\\
    Gemini-2.5-Pro &75.90 &44.05 &59.98	&\light{99.68}	&54.77	&29.32	&61.26	&52.11	&87.89	&38.11	&53.18	&57.82	&59.50 &4\\
    GPT-4.1 & 74.38  &46.59	&60.49	&97.73	&53.18	&35.23	&\light{62.05}	&49.73	&\light{96.82}	&\light{58.64}	&52.52	& 64.43	&62.53 & 2 \\
    Grok-4 &74.84 &49.27 &62.06	&95.68	&38.18	&\light{41.39}	&58.42	&\light{55.40}	&86.20	&55.68	&\light{69.32}	&\light{66.65}	&\light{62.80} &1\\
    Llama-3.2-90B-Vision &66.16 &41.77 &53.97	&90.50	&37.26	&33.41	&53.72	&48.41	&88.00	&45.23	&39.07	&55.18	&54.38 &7\\
    Pixtral-12B &64.64 &41.38 &53.01	&90.76	&44.45	&22.73	&52.65	&45.39	&84.76	&27.03	&43.89	&50.27	&51.80 &10\\
    O3 &\light{76.61} &\light{53.02} &\light{64.82}	&98.52	&49.98	&28.39	&58.96	&51.72	&83.86	&47.40	&58.18	&60.29	&61.25 &3\\
      \midrule
    \end{tabular}
    }
    \vspace{-0.2cm}
\end{table*}

\begin{figure}[t]
    \centering
    \vspace{-0.3cm}
    \includegraphics[width=1.0\linewidth]{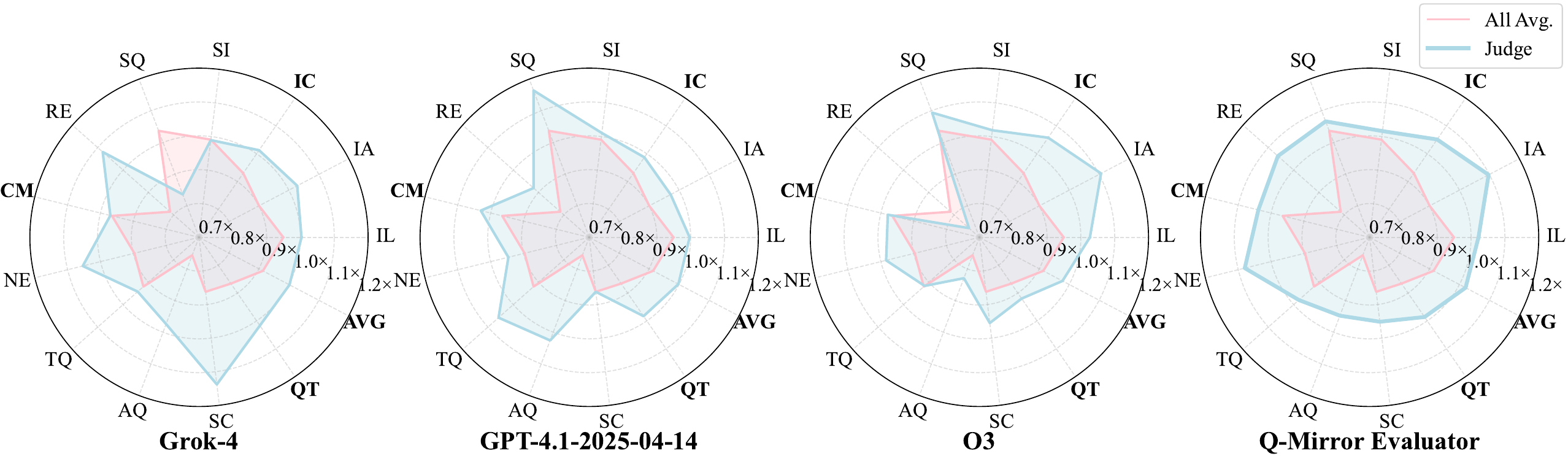}
    \caption{Performance comparison for the top-3 judge models and Q-Mirror \textit{Evaluator} (judge ensemble group).  In each chart, the \textbf{red line} shows the ratio between the overall average of all ten judges and the average of the top three ($\text{Avg}{10} / \text{Avg}{\text{Top-3}}$). The \textbf{blue line} indicates the relative performance of the current judge (or ensemble) compared with the top-three average ($\text{Score}{\text{current}} /\text{Avg}{\text{Top-3}}$). }
    \label{fig:radar}
    \vspace{-0.2cm}
\end{figure}

\subsection{Findings}
\label{sec:generation_benchmark}

\paragraph{Finding for MMQA Generation.}

As shown in Table~\ref{tab:generation}, the \textbf{GPT Family} achieves the highest overall quality with an average score of 78.90, which highlights its strong capability for scientific T2I tasks and underscores its potential as the base module for Q-Mirror. Secondly, a clear trend emerges across all model families: performance on Q-Mirror-Grad consistently surpasses that on Q-Mirror-Expert. This gap might indicate that current models handle structured knowledge with standard visual patterns more effectively than expert-level concepts that lack canonical visualizations. Thirdly, a consistent pattern is observed across all model families: they achieve near-perfect scores on the SI dimension (i.e., whether the question can be solved from the image alone). This indicates that the generated scientific images are reasonably self-contained without revealing the answer directly. In contrast, their performance on the SC dimension (i.e., whether the image is plausible and scientifically sound) is consistently poor. This stark disparity highlights a significant limitation of current T2I models, particularly in rendering legible text and ensuring scientific plausibility, indicating a substantial gap that must be bridged for practical application.

\paragraph{Findings for MMQA Evaluation.}

\begin{figure*}[t]
  \centering
  \subfloat[DeepSeek-VL2]{\includegraphics[width=0.32\linewidth]{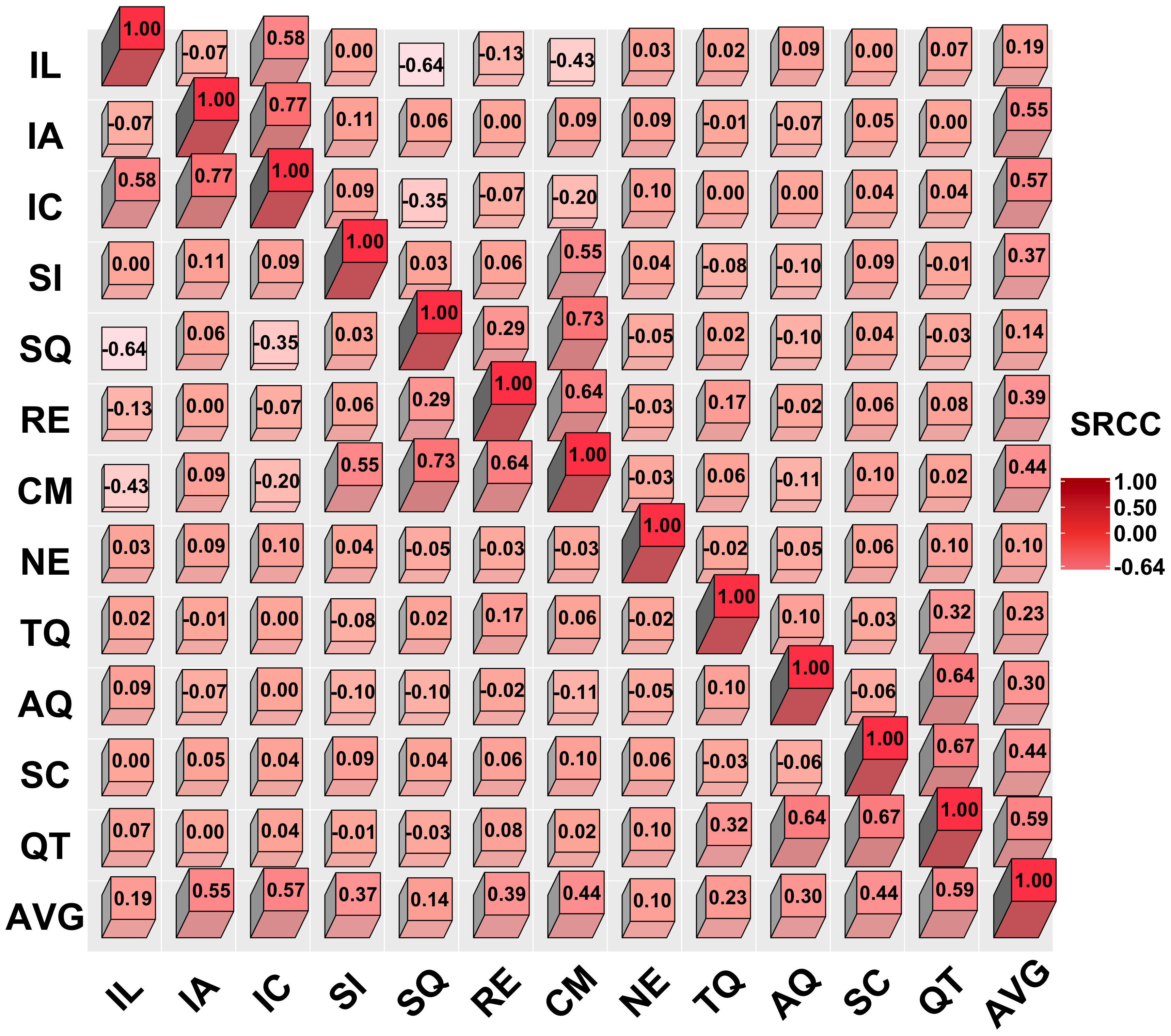}\label{fig:deepseek}}
  \hfill
  \subfloat[Claude-Sonnet-4]{\includegraphics[width=0.32\linewidth]{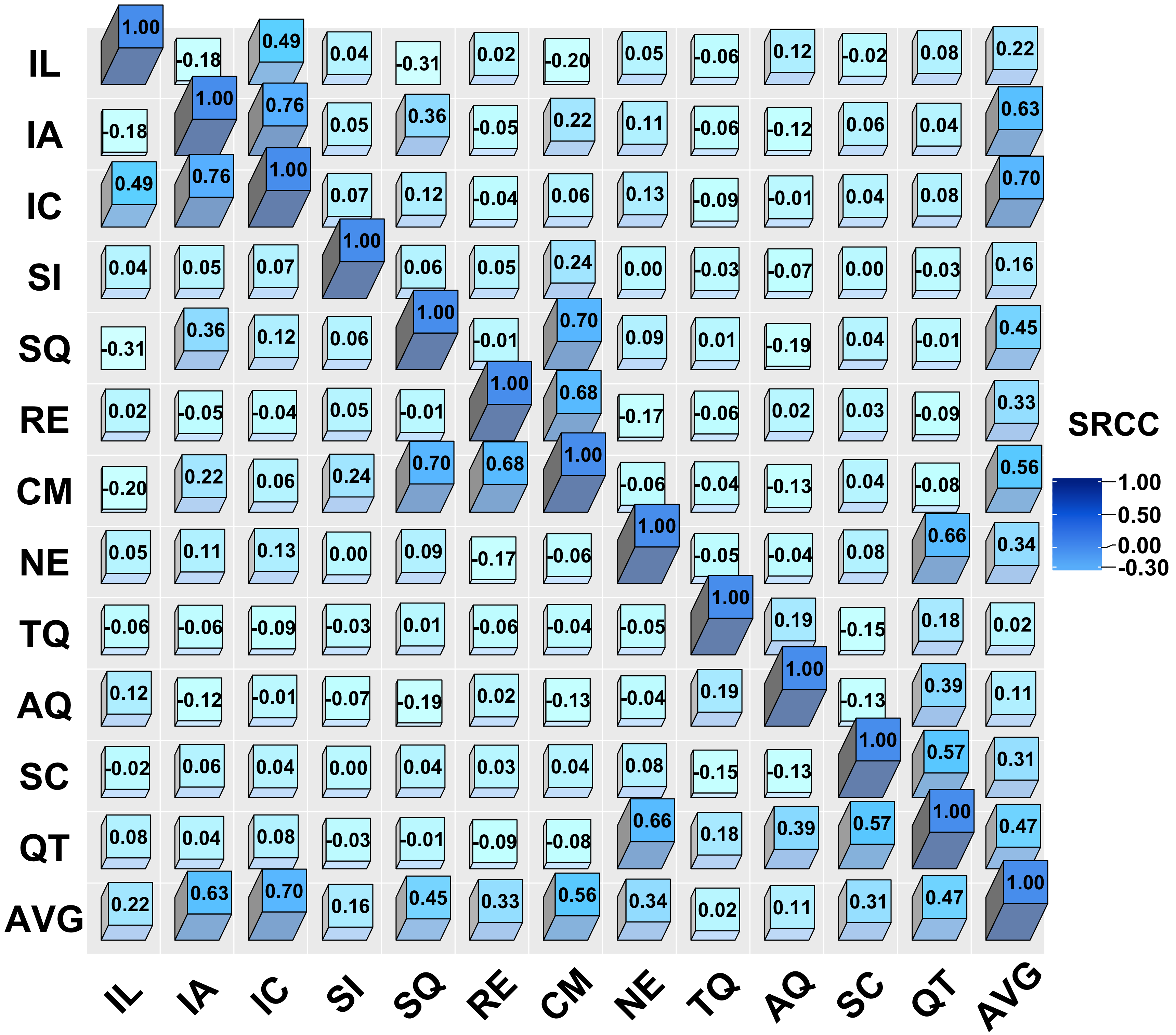}\label{fig:claude}}
  \hfill
  \subfloat[Qwen2-VL-72B-Instruct]{\includegraphics[width=0.32\linewidth]{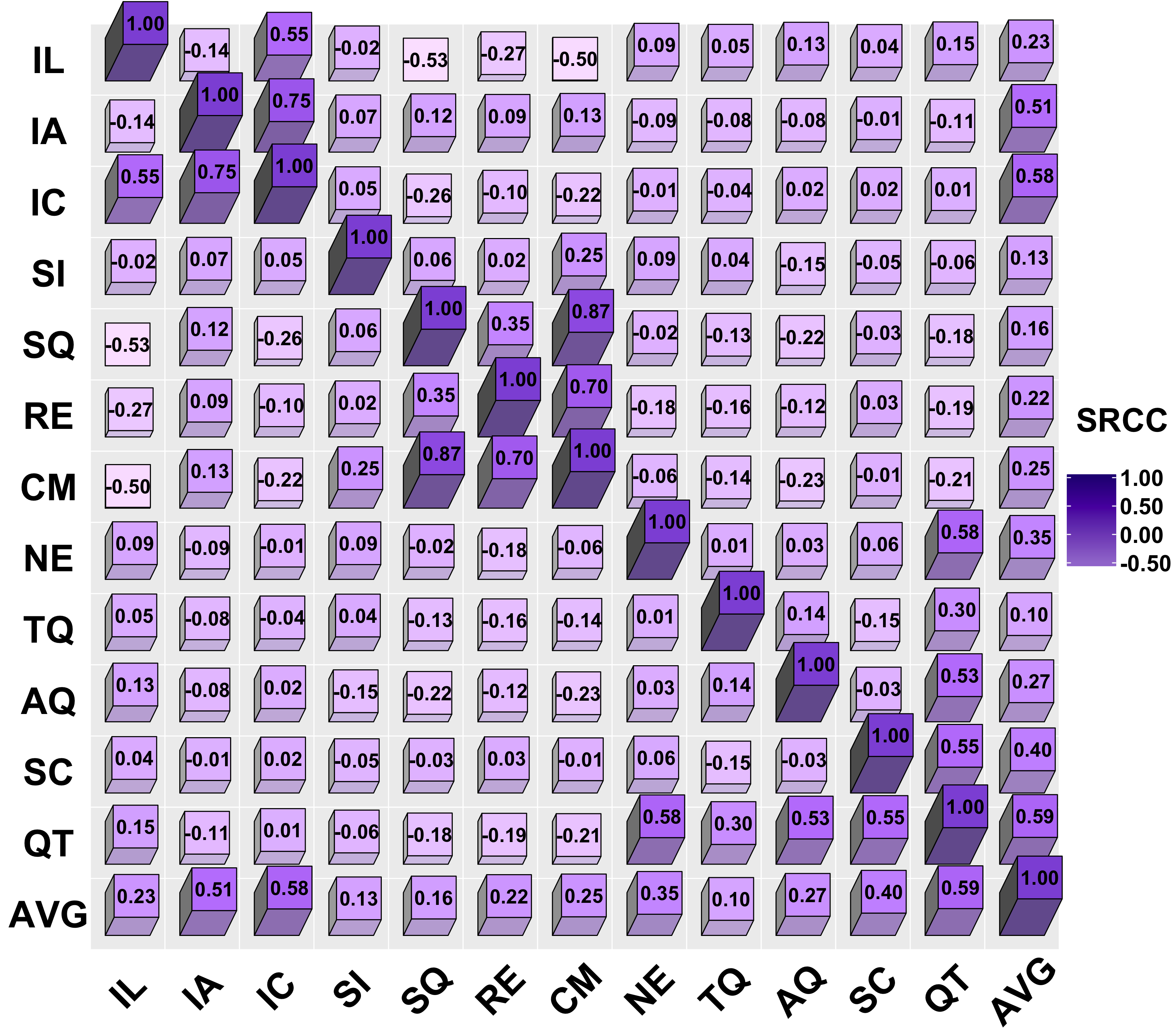}\label{fig:qwen}}
   \vspace{-0.2cm}
  \caption{Dimensional SRCC correlations of MMQA evaluation, following the setting of  Redundancy Principle~\citep{Redundancy}. Higher and darker bars indicate stronger agreement between the corresponding dimensions.}
  \label{fig:SRCC}
  \vspace{-0.3cm}
\end{figure*}

Table~\ref{tab:judge} reveals two key observations. First, judge performance is clearly stratified: a top tier led by \textit{Grok-4} achieves the highest agreement with human annotations (62.80\%), whereas several weaker models fall below 55\%. Second, task difficulty varies with the abstraction level. All judges align more closely with humans on Q-Mirror-Grad than on Q-Mirror-Expert. For instance, \textit{Claude-Sonnet-4} improves from 58.98\% on Expert to 71.00\% on Grad, reflecting that graduate-level questions admit more objective criteria, while expert-level questions involve subjective reasoning, which increase divergence. Third, as illustrated in Figure~\ref{fig:SRCC}, the correlations among the major dimensions (e.g., IC, CM, QT) are low, which validates the overall non-redundancy of the benchmark design~\citep{Redundancy}. Furthermore, this result indirectly affirms the effectiveness of the proposed quality principles.

\subsection{Performance of Q-Mirror}
\label{sec:agent_performance}
The findings above highlight the risk of relying on a single judge, as its biases would directly affect evaluation, especially for complex scientific content. To mitigate this, we adopt a Judge Ensemble for Q-Mirror composed of the top three models \textit{Grok-4}, \textit{GPT-4.1-2025-04-14}, and \textit{O3} (\textit{judge performance shown in Figure \ref{fig:radar}}), which aggregates diverse perspectives, reduces model-specific bias. In total, the judge ensemble provides more stable and reliable feedback across dimensions, mitigates single-model biases, and guarantees  closer alignment with human judgment. 

In addition, Table~\ref{tab:generation} demonstrates that the Q-Mirror agent consistently outperforms all baseline families, improving the overall average score from $78.90$ to $85.22$ and raising the pass rate from 72\% to 95\% (\textit{definition in~\S\ref{sec:setup}}). This demonstrates that Q-Mirror not only enhances overall quality but also ensures greater stability and consistency in producing high-quality MMQAs. Further analysis reveals critical insights: While models excel at solvability from images alone, they consistently struggle with scientific plausibility, underscoring the difficulty of rendering domain-accurate visuals. Moreover, the greater improvements on expert-level tasks suggest that Q-Mirror is particularly valuable for challenging, less standardized scientific problems.

\section{Conclusion}

This work addresses the critical scarcity of multi-modal scientific QA pairs by introducing Q-Mirror, a systematic framework for converting TQAs into high-quality MMQAs. Our contributions are threefold. First, we establish a \textbf{transformation pipeline and comprehensive quality rubric} that define principles for effective SMQ-to-MMQA transformation. Second, we introduce \textbf{two extensive benchmarks} for MMQA generation and evaluation, revealing distinct strengths and weaknesses in current state-of-the-art models. Third, we implement the \textbf{Q-Mirror}, an autonomous system that integrates generation and judgment through iterative refinement. The agent achieves an average quality score of 85.22 and a 95\% pass rate, demonstrating the efficacy of our feedback-driven architecture. By unlocking the multi-modal potential of text-based corpora, Q-Mirror provides a scalable and cost-effective approach to advance scientific AI evaluation and support the development of next-generation reasoning models.

\bibliography{iclr2026_conference}

\begin{thebibliography}{55}
\providecommand{\natexlab}[1]{#1}
\providecommand{\url}[1]{\texttt{#1}}
\expandafter\ifx\csname urlstyle\endcsname\relax
  \providecommand{\doi}[1]{doi: #1}\else
  \providecommand{\doi}{doi: \begingroup \urlstyle{rm}\Url}\fi

\bibitem[Agrawal et~al.(2024)Agrawal, Antoniak, Hanna, Bout, Chaplot, Chudnovsky, Costa, De~Monicault, Garg, Gervet, et~al.]{agrawal2024pixtral}
Pravesh Agrawal, Szymon Antoniak, Emma~Bou Hanna, Baptiste Bout, Devendra Chaplot, Jessica Chudnovsky, Diogo Costa, Baudouin De~Monicault, Saurabh Garg, Theophile Gervet, et~al.
\newblock Pixtral 12b.
\newblock \emph{arXiv preprint arXiv:2410.07073}, 2024.

\bibitem[Anthropic(2025)]{anthropic_claude4_2025}
Anthropic.
\newblock Introducing claude 4.
\newblock \url{https://www.anthropic.com/news/claude-4}, May 2025.

\bibitem[Bai et~al.(2023)Bai, Bai, Chu, Cui, Dang, Deng, Fan, Ge, Han, Huang, et~al.]{bai2023qwen}
Jinze Bai, Shuai Bai, Yunfei Chu, Zeyu Cui, Kai Dang, Xiaodong Deng, Yang Fan, Wenbin Ge, Yu~Han, Fei Huang, et~al.
\newblock Qwen technical report.
\newblock \emph{arXiv preprint arXiv:2309.16609}, 2023.

\bibitem[Bai et~al.(2025)Bai, Cai, Cao, Cao, Cao, Chen, Chen, Chen, Chen, Chen, et~al.]{Intern-S1}
Lei Bai, Zhongrui Cai, Yuhang Cao, Maosong Cao, Weihan Cao, Chiyu Chen, Haojiong Chen, Kai Chen, Pengcheng Chen, Ying Chen, et~al.
\newblock Intern-s1: A scientific multimodal foundation model, 2025.
\newblock URL \url{https://arxiv.org/abs/2508.15763}.

\bibitem[ByteDance(2024)]{bytedance_doubao_2024}
ByteDance.
\newblock Doubao model series.
\newblock \url{https://www.doubao.com/}, 2024.

\bibitem[Chan et~al.(2023)Chan, Chen, Su, Yu, Xue, Zhang, Fu, and Liu]{ChatEval}
Chi-Min Chan, Weize Chen, Yusheng Su, Jianxuan Yu, Wei Xue, Shanghang Zhang, Jie Fu, and Zhiyuan Liu.
\newblock Chateval: Towards better llm-based evaluators through multi-agent debate, 2023.
\newblock URL \url{https://arxiv.org/abs/2308.07201}.

\bibitem[Cho et~al.(2025)Cho, Qin, Liu, Choi, Lee, and Kim]{Geometry}
Seunghyuk Cho, Zhenyue Qin, Yang Liu, Youngbin Choi, Seungbeom Lee, and Dongwoo Kim.
\newblock Plane geometry problem solving with multi-modal reasoning: A survey, 2025.
\newblock URL \url{https://arxiv.org/abs/2505.14340}.

\bibitem[Gao et~al.(2025)Gao, Gong, Guo, Hou, Lai, Li, Li, Lian, Liao, Liu, et~al.]{gao2025seedream}
Yu~Gao, Lixue Gong, Qiushan Guo, Xiaoxia Hou, Zhichao Lai, Fanshi Li, Liang Li, Xiaochen Lian, Chao Liao, Liyang Liu, et~al.
\newblock Seedream 3.0 technical report.
\newblock \emph{arXiv preprint arXiv:2504.11346}, 2025.

\bibitem[Grattafiori et~al.(2024)Grattafiori, Dubey, Jauhri, Pandey, Kadian, Al-Dahle, Letman, Mathur, Schelten, Vaughan, et~al.]{grattafiori2024llama}
Aaron Grattafiori, Abhimanyu Dubey, Abhinav Jauhri, Abhinav Pandey, Abhishek Kadian, Ahmad Al-Dahle, Aiesha Letman, Akhil Mathur, Alan Schelten, Alex Vaughan, et~al.
\newblock The llama 3 herd of models.
\newblock \emph{arXiv preprint arXiv:2407.21783}, 2024.

\bibitem[Hendrycks et~al.(2020)Hendrycks, Burns, Basart, Zou, Mazeika, Song, and Steinhardt]{hendrycks2020measuring}
Dan Hendrycks, Collin Burns, Steven Basart, Andy Zou, Mantas Mazeika, Dawn Song, and Jacob Steinhardt.
\newblock Measuring massive multitask language understanding.
\newblock \emph{arXiv preprint arXiv:2009.03300}, 2020.

\bibitem[Hendrycks et~al.(2021)Hendrycks, Burns, Kadavath, Arora, Basart, Tang, Song, and Steinhardt]{MATH}
Dan Hendrycks, Collin Burns, Saurav Kadavath, Akul Arora, Steven Basart, Eric Tang, Dawn Song, and Jacob Steinhardt.
\newblock Measuring mathematical problem solving with the math dataset, 2021.
\newblock URL \url{https://arxiv.org/abs/2103.03874}.

\bibitem[Hu et~al.(2025)Hu, Ma, Li, Xu, Wu, Hu, Li, Zhuang, Liu, Lu, Chen, et~al.]{SLLM}
Ming Hu, Chenglong Ma, Wei Li, Wanghan Xu, Jiamin Wu, Jucheng Hu, Tianbin Li, Guohang Zhuang, Jiaqi Liu, Yingzhou Lu, Ying Chen, et~al.
\newblock A survey of scientific large language models: From data foundations to agent frontiers, 2025.
\newblock URL \url{https://arxiv.org/abs/2508.21148}.

\bibitem[Huang et~al.(2023)Huang, Bai, Zhu, Zhang, Zhang, Su, Liu, Lv, Zhang, Fu, et~al.]{huang2023c}
Yuzhen Huang, Yuzhuo Bai, Zhihao Zhu, Junlei Zhang, Jinghan Zhang, Tangjun Su, Junteng Liu, Chuancheng Lv, Yikai Zhang, Yao Fu, et~al.
\newblock C-eval: A multi-level multi-discipline chinese evaluation suite for foundation models.
\newblock \emph{Advances in Neural Information Processing Systems}, 36:\penalty0 62991--63010, 2023.

\bibitem[Joshi et~al.(2017)Joshi, Choi, Weld, and Zettlemoyer]{DBLP:journals/corr/JoshiCWZ17}
Mandar Joshi, Eunsol Choi, Daniel~S. Weld, and Luke Zettlemoyer.
\newblock Triviaqa: {A} large scale distantly supervised challenge dataset for reading comprehension.
\newblock \emph{CoRR}, abs/1705.03551, 2017.
\newblock URL \url{http://arxiv.org/abs/1705.03551}.

\bibitem[Kembhavi et~al.(2016)Kembhavi, Salvato, Kolve, Seo, Hajishirzi, and Farhadi]{kembhavi2016diagram}
Aniruddha Kembhavi, Mike Salvato, Eric Kolve, Minjoon Seo, Hannaneh Hajishirzi, and Ali Farhadi.
\newblock A diagram is worth a dozen images.
\newblock In \emph{European conference on computer vision}, pp.\  235--251. Springer, 2016.

\bibitem[K{\"u}chemann et~al.(2025)K{\"u}chemann, Avila, Dinc, Hortmann, Revenga, Ruf, Stausberg, Steinert, Fischer, Fischer, Kasneci, Kasneci, Kuhr, Kutyniok, Malone, Sailer, Schmidt, Stadler, Weller, and Kuhn]{Kuchemann2025}
Stefan K{\"u}chemann, Karina~E. Avila, Yavuz Dinc, Chiara Hortmann, Natalia Revenga, Verena Ruf, Niklas Stausberg, Steffen Steinert, Frank Fischer, Martin Fischer, Enkelejda Kasneci, Gjergji Kasneci, Thomas Kuhr, Gitta Kutyniok, Sarah Malone, Michael Sailer, Albrecht Schmidt, Matthias Stadler, Jochen Weller, and Jochen Kuhn.
\newblock On opportunities and challenges of large multimodal foundation models in education.
\newblock \emph{npj Science of Learning}, 10\penalty0 (1):\penalty0 11, 2025.
\newblock \doi{10.1038/s41539-025-00301-w}.
\newblock URL \url{https://doi.org/10.1038/s41539-025-00301-w}.

\bibitem[Li et~al.(2024{\natexlab{a}})Li, Zhang, Koto, Yang, Zhao, Gong, Duan, and Baldwin]{CMMLU}
Haonan Li, Yixuan Zhang, Fajri Koto, Yifei Yang, Hai Zhao, Yeyun Gong, Nan Duan, and Timothy Baldwin.
\newblock Cmmlu: Measuring massive multitask language understanding in chinese, 2024{\natexlab{a}}.
\newblock URL \url{https://arxiv.org/abs/2306.09212}.

\bibitem[Li et~al.(2025)]{li2025chemvlm}
Junxian Li et~al.
\newblock Chemvlm: Exploring the power of multimodal large language models in chemistry area.
\newblock In \emph{Proceedings of the AAAI Conference on Artificial Intelligence}, volume~39, 2025.

\bibitem[Li et~al.(2024{\natexlab{b}})Li, Chen, Shi, Xiao, and Chen]{multimodalbench}
Lin Li, Guikun Chen, Hanrong Shi, Jun Xiao, and Long Chen.
\newblock A survey on multimodal benchmarks: In the era of large ai models, 2024{\natexlab{b}}.
\newblock URL \url{https://arxiv.org/abs/2409.18142}.

\bibitem[Liu et~al.(2025)Liu, Zhou, Guo, Shareghi, Vulić, Korhonen, and Collier]{liu2025aligninghumanjudgementrole}
Yinhong Liu, Han Zhou, Zhijiang Guo, Ehsan Shareghi, Ivan Vulić, Anna Korhonen, and Nigel Collier.
\newblock Aligning with human judgement: The role of pairwise preference in large language model evaluators, 2025.
\newblock URL \url{https://arxiv.org/abs/2403.16950}.

\bibitem[Liu et~al.(2023)Liu, Tang, Shi, Zhang, Li, Shrivastava, and Wilson]{data_aug}
Zichang Liu, Zhiqiang Tang, Xingjian Shi, Aston Zhang, Mu~Li, Anshumali Shrivastava, and Andrew~Gordon Wilson.
\newblock Learning multimodal data augmentation in feature space, 2023.
\newblock URL \url{https://arxiv.org/abs/2212.14453}.

\bibitem[Lu et~al.(2022)Lu, Mishra, Xia, Qiu, Chang, Zhu, Tafjord, Clark, and Kalyan]{lu2022learn}
Pan Lu, Swaroop Mishra, Tanglin Xia, Liang Qiu, Kai-Wei Chang, Song-Chun Zhu, Oyvind Tafjord, Peter Clark, and Ashwin Kalyan.
\newblock Learn to explain: Multimodal reasoning via thought chains for science question answering.
\newblock \emph{Advances in Neural Information Processing Systems}, 35:\penalty0 2507--2521, 2022.

\bibitem[Lu et~al.(2023)Lu, Bansal, Xia, Liu, Li, Hajishirzi, Cheng, Chang, Galley, and Gao]{lu2023mathvista}
Pan Lu, Hritik Bansal, Tony Xia, Jiacheng Liu, Chunyuan Li, Hannaneh Hajishirzi, Hao Cheng, Kai-Wei Chang, Michel Galley, and Jianfeng Gao.
\newblock Mathvista: Evaluating mathematical reasoning of foundation models in visual contexts.
\newblock \emph{arXiv preprint arXiv:2310.02255}, 2023.

\bibitem[Ning et~al.(2023)Ning, Wang, Huang, and Huang]{Symbolic}
Maizhen Ning, Qiu-Feng Wang, Kaizhu Huang, and Xiaowei Huang.
\newblock A symbolic character-aware model for solving geometry problems, 2023.
\newblock URL \url{https://arxiv.org/abs/2308.02823}.

\bibitem[OpenAI(2025{\natexlab{a}})]{openai_gpt41_2025}
OpenAI.
\newblock Introducing gpt-4.1 in the api.
\newblock \url{https://openai.com/index/gpt-4-1/}, 4 2025{\natexlab{a}}.

\bibitem[OpenAI(2025{\natexlab{b}})]{openai_imagegen_2025}
OpenAI.
\newblock Image generation: Learn how to generate or edit images.
\newblock \url{https://platform.openai.com/docs/guides/image-generation?image-generation-model=gpt-image-1}, 5 2025{\natexlab{b}}.

\bibitem[OpenAI(2025{\natexlab{c}})]{openai_o3_o4mini_2025}
OpenAI.
\newblock Introducing openai o3 and o4-mini.
\newblock \url{https://openai.com/zh-Hans-CN/index/introducing-o3-and-o4-mini}, 2025{\natexlab{c}}.

\bibitem[Phan et~al.(2025)Phan, Gatti, Han, Li, Hu, Zhang, Zhang, Shaaban, Ling, et~al.]{HLE}
Long Phan, Alice Gatti, Ziwen Han, Nathaniel Li, Josephina Hu, Hugh Zhang, Chen Bo~Calvin Zhang, Mohamed Shaaban, John Ling, et~al.
\newblock Humanity's last exam, 2025.
\newblock URL \url{https://arxiv.org/abs/2501.14249}.

\bibitem[Qin et~al.(2025)Qin, Chen, Wang, Wu, Chen, Cheng, Zhao, Xiao, Dong, Long, Pan, Wu, Li, Zhou, Xiong, and Zhu]{SciHorizon}
Chuan Qin, Xin Chen, Chengrui Wang, Pengmin Wu, Xi~Chen, Yihang Cheng, Jingyi Zhao, Meng Xiao, Xiangchao Dong, Qingqing Long, Boya Pan, Han Wu, Chengzan Li, Yuanchun Zhou, Hui Xiong, and Hengshu Zhu.
\newblock Scihorizon: Benchmarking ai-for-science readiness from scientific data to large language models, 2025.
\newblock URL \url{https://arxiv.org/abs/2503.13503}.

\bibitem[Rein et~al.(2024)Rein, Hou, Stickland, Petty, Pang, Dirani, Michael, and Bowman]{rein2024gpqa}
David Rein, Betty~Li Hou, Asa~Cooper Stickland, Jackson Petty, Richard~Yuanzhe Pang, Julien Dirani, Julian Michael, and Samuel~R Bowman.
\newblock Gpqa: A graduate-level google-proof q\&a benchmark.
\newblock In \emph{First Conference on Language Modeling}, 2024.

\bibitem[Sun et~al.(2024)Sun, Han, Zhao, Ma, Shen, Chen, Chen, and Yu]{SciEval}
Liangtai Sun, Yang Han, Zihan Zhao, Da~Ma, Zhennan Shen, Baocai Chen, Lu~Chen, and Kai Yu.
\newblock Scieval: A multi-level large language model evaluation benchmark for scientific research, 2024.
\newblock URL \url{https://arxiv.org/abs/2308.13149}.

\bibitem[Tan et~al.(2025)Tan, Zhuang, Montgomery, Tang, Cuadron, Wang, Popa, and Stoica]{JudgeBench}
Sijun Tan, Siyuan Zhuang, Kyle Montgomery, William~Y. Tang, Alejandro Cuadron, Chenguang Wang, Raluca~Ada Popa, and Ion Stoica.
\newblock Judgebench: A benchmark for evaluating llm-based judges, 2025.
\newblock URL \url{https://arxiv.org/abs/2410.12784}.

\bibitem[Team et~al.(2023)]{team2023gemini}
Gemini Team et~al.
\newblock Gemini: a family of highly capable multimodal models.
\newblock \emph{arXiv preprint arXiv:2312.11805}, 2023.

\bibitem[Team et~al.(2025)Team, Du, Yao, Ma, Wang, Zheng, Zhu, Liu, Liang, et~al.]{SuperGPQA}
P~Team, Xinrun Du, Yifan Yao, Kaijing Ma, Bingli Wang, Tianyu Zheng, King Zhu, Minghao Liu, Yiming Liang, et~al.
\newblock Supergpqa: Scaling llm evaluation across 285 graduate disciplines, 2025.
\newblock URL \url{https://arxiv.org/abs/2502.14739}.

\bibitem[Villalobos et~al.(2022)Villalobos, Ho, Sevilla, Besiroglu, Heim, and Hobbhahn]{villalobos2022will}
Pablo Villalobos, Anson Ho, Jaime Sevilla, Tamay Besiroglu, Lennart Heim, and Marius Hobbhahn.
\newblock Will we run out of data? limits of llm scaling based on human-generated data.
\newblock \emph{arXiv preprint arXiv:2211.04325}, 2022.

\bibitem[Wadhwa et~al.(2025)Wadhwa, Chen, Li, and Durrett]{wadhwa2025usingnaturallanguageexplanations}
Manya Wadhwa, Jifan Chen, Junyi~Jessy Li, and Greg Durrett.
\newblock Using natural language explanations to rescale human judgments, 2025.
\newblock URL \url{https://arxiv.org/abs/2305.14770}.

\bibitem[Wan et~al.(2025)Wan, Wang, Ai, Wen, Mao, Xie, Chen, Yu, Zhao, Yang, et~al.]{wan2025wan}
Team Wan, Ang Wang, Baole Ai, Bin Wen, Chaojie Mao, Chen-Wei Xie, Di~Chen, Feiwu Yu, Haiming Zhao, Jianxiao Yang, et~al.
\newblock Wan: Open and advanced large-scale video generative models.
\newblock \emph{arXiv preprint arXiv:2503.20314}, 2025.

\bibitem[Wang et~al.(2025{\natexlab{a}})Wang, Li, Wu, Liang, Guo, Li, Duan, Zhang, and Zhai]{wang2025affordance}
Junying Wang, Wenzhe Li, Yalun Wu, Yingji Liang, Yijin Guo, Chunyi Li, Haodong Duan, Zicheng Zhang, and Guangtao Zhai.
\newblock Affordance benchmark for mllms.
\newblock \emph{arXiv preprint arXiv:2506.00893}, 2025{\natexlab{a}}.

\bibitem[Wang et~al.(2025{\natexlab{b}})Wang, Zhang, and Yuan]{Adv-CPG}
Junying Wang, Hongyuan Zhang, and Yuan Yuan.
\newblock Adv-cpg: A customized portrait generation framework with facial adversarial attacks.
\newblock In \emph{Proceedings of the Computer Vision and Pattern Recognition Conference (CVPR)}, pp.\  21001--21010, June 2025{\natexlab{b}}.

\bibitem[Wang et~al.(2025{\natexlab{c}})Wang, Zhang, Guo, Wen, Shen, Liang, Wu, Li, Li, Chen, Jia, and Zhai]{eese}
Junying Wang, Zicheng Zhang, Yijin Guo, Farong Wen, Ye~Shen, Yingji Liang, Yalun Wu, Wenzhe Li, Chunyi Li, Zijian Chen, Qi~Jia, and Guangtao Zhai.
\newblock The ever-evolving science exam, 2025{\natexlab{c}}.
\newblock URL \url{https://arxiv.org/abs/2507.16514}.

\bibitem[Wang et~al.(2024{\natexlab{a}})Wang, Hu, He, Xu, Liu, Liu, and Shen]{T-SciQ}
Lei Wang, Yi~Hu, Jiabang He, Xing Xu, Ning Liu, Hui Liu, and Heng~Tao Shen.
\newblock T-sciq: Teaching multimodal chain-of-thought reasoning via large language model signals for science question answering.
\newblock \emph{Proceedings of the AAAI Conference on Artificial Intelligence}, 38\penalty0 (17):\penalty0 19162--19170, Mar. 2024{\natexlab{a}}.

\bibitem[Wang et~al.(2024{\natexlab{b}})Wang, Bai, Tan, Wang, Fan, Bai, Chen, Liu, Wang, Ge, et~al.]{wang2024qwen2}
Peng Wang, Shuai Bai, Sinan Tan, Shijie Wang, Zhihao Fan, Jinze Bai, Keqin Chen, Xuejing Liu, Jialin Wang, Wenbin Ge, et~al.
\newblock Qwen2-vl: Enhancing vision-language model's perception of the world at any resolution.
\newblock \emph{arXiv preprint arXiv:2409.12191}, 2024{\natexlab{b}}.

\bibitem[Wang et~al.(2025{\natexlab{d}})Wang, Wu, Zhang, Yan, Liu, Luo, and Fei]{MMCOT}
Yaoting Wang, Shengqiong Wu, Yuecheng Zhang, Shuicheng Yan, Ziwei Liu, Jiebo Luo, and Hao Fei.
\newblock Multimodal chain-of-thought reasoning: A comprehensive survey, 2025{\natexlab{d}}.
\newblock URL \url{https://arxiv.org/abs/2503.12605}.

\bibitem[Wang et~al.(2024{\natexlab{c}})Wang, Ma, Zhang, Ni, Chandra, Guo, Ren, Arulraj, He, Jiang, Li, Ku, Wang, Zhuang, Fan, Yue, and Chen]{MMLU-Pro}
Yubo Wang, Xueguang Ma, Ge~Zhang, Yuansheng Ni, Abhranil Chandra, Shiguang Guo, Weiming Ren, Aaran Arulraj, Xuan He, Ziyan Jiang, Tianle Li, Max Ku, Kai Wang, Alex Zhuang, Rongqi Fan, Xiang Yue, and Wenhu Chen.
\newblock Mmlu-pro: A more robust and challenging multi-task language understanding benchmark, 2024{\natexlab{c}}.
\newblock URL \url{https://arxiv.org/abs/2406.01574}.

\bibitem[Wang et~al.(2025{\natexlab{e}})Wang, Wang, Liu, Wang, Fu, Lu, Aggarwal, Pei, and Zhou]{aug_survey}
Zaitian Wang, Pengfei Wang, Kunpeng Liu, Pengyang Wang, Yanjie Fu, Chang-Tien Lu, Charu~C. Aggarwal, Jian Pei, and Yuanchun Zhou.
\newblock A comprehensive survey on data augmentation, 2025{\natexlab{e}}.
\newblock URL \url{https://arxiv.org/abs/2405.09591}.

\bibitem[Wu et~al.(2025)Wu, Li, Zhou, Lin, Gao, Yan, Yin, Bai, Xu, Chen, et~al.]{wu2025qwen}
Chenfei Wu, Jiahao Li, Jingren Zhou, Junyang Lin, Kaiyuan Gao, Kun Yan, Sheng-ming Yin, Shuai Bai, Xiao Xu, Yilei Chen, et~al.
\newblock Qwen-image technical report.
\newblock \emph{arXiv preprint arXiv:2508.02324}, 2025.

\bibitem[Wu et~al.(2024)Wu, Chen, Pan, Liu, Liu, Dai, Gao, Ma, Wu, Wang, Xie, Wu, Hu, Wang, Sun, Li, Piao, Guan, Liu, Xie, You, Dong, Yu, Zhang, Zhao, Wang, and Ruan]{deepseekvl2}
Zhiyu Wu, Xiaokang Chen, Zizheng Pan, Xingchao Liu, Wen Liu, Damai Dai, Huazuo Gao, Yiyang Ma, Chengyue Wu, Bingxuan Wang, Zhenda Xie, Yu~Wu, Kai Hu, Jiawei Wang, Yaofeng Sun, Yukun Li, Yishi Piao, Kang Guan, Aixin Liu, Xin Xie, Yuxiang You, Kai Dong, Xingkai Yu, Haowei Zhang, Liang Zhao, Yisong Wang, and Chong Ruan.
\newblock Deepseek-vl2: Mixture-of-experts vision-language models for advanced multimodal understanding, 2024.
\newblock URL \url{https://arxiv.org/abs/2412.10302}.

\bibitem[xAI(2025{\natexlab{a}})]{xai_api_image_2025}
xAI.
\newblock xai api guide: Image generations.
\newblock \url{https://docs.x.ai/docs/guides/image-generations}, 2025{\natexlab{a}}.

\bibitem[xAI(2025{\natexlab{b}})]{xai_grok4_2025}
xAI.
\newblock Grok 4.
\newblock \url{https://x.ai/news/grok-4}, July 2025{\natexlab{b}}.

\bibitem[Yin et~al.(2024)Yin, Fu, Zhao, Li, Sun, Xu, and Chen]{Yin_2024}
Shukang Yin, Chaoyou Fu, Sirui Zhao, Ke~Li, Xing Sun, Tong Xu, and Enhong Chen.
\newblock A survey on multimodal large language models.
\newblock \emph{National Science Review}, 11\penalty0 (12), November 2024.
\newblock ISSN 2053-714X.
\newblock \doi{10.1093/nsr/nwae403}.
\newblock URL \url{http://dx.doi.org/10.1093/nsr/nwae403}.

\bibitem[Yue et~al.(2024)Yue, Ni, Zhang, Zheng, Liu, Zhang, Stevens, Jiang, Ren, Sun, et~al.]{yue2024mmmu}
Xiang Yue, Yuansheng Ni, Kai Zhang, Tianyu Zheng, Ruoqi Liu, Ge~Zhang, Samuel Stevens, Dongfu Jiang, Weiming Ren, Yuxuan Sun, et~al.
\newblock Mmmu: A massive multi-discipline multimodal understanding and reasoning benchmark for expert agi.
\newblock In \emph{Proceedings of the IEEE/CVF Conference on Computer Vision and Pattern Recognition}, pp.\  9556--9567, 2024.

\bibitem[Zhang et~al.(2025{\natexlab{a}})Zhang, Yao, Pi, Liang, and Fung]{VLM2-Bench}
Jianshu Zhang, Dongyu Yao, Renjie Pi, Paul~Pu Liang, and Yi~R. Fung.
\newblock Vlm2-bench: A closer look at how well vlms implicitly link explicit matching visual cues, 2025{\natexlab{a}}.
\newblock URL \url{https://arxiv.org/abs/2502.12084}.

\bibitem[Zhang et~al.(2025{\natexlab{b}})Zhang, Wang, Wen, Guo, Zhao, Fang, Ding, Jia, Xiao, Shen, Zheng, Zhu, Wu, Jiao, Sun, Chen, Zhang, Fu, Cao, Hu, Zhou, Zhou, Cao, Zhou, Cao, Li, Zhou, Tian, Zhu, Li, Wu, Liu, He, Zhou, Liu, Zhang, Wang, Duan, Zhou, Min, Jia, Zhou, Zhang, Cao, Yang, Yu, Zhang, Duan, and Zhai]{zhang2025lmmsurvey}
Zicheng Zhang, Junying Wang, Farong Wen, Yijin Guo, Xiangyu Zhao, Xinyu Fang, Shengyuan Ding, Ziheng Jia, Jiahao Xiao, Ye~Shen, Yushuo Zheng, Xiaorong Zhu, Yalun Wu, Ziheng Jiao, Wei Sun, Zijian Chen, Kaiwei Zhang, Kang Fu, Yuqin Cao, Ming Hu, Yue Zhou, Xuemei Zhou, Juntai Cao, Wei Zhou, Jinyu Cao, Ronghui Li, Donghao Zhou, Yuan Tian, Xiangyang Zhu, Chunyi Li, Haoning Wu, Xiaohong Liu, Junjun He, Yu~Zhou, Hui Liu, Lin Zhang, Zesheng Wang, Huiyu Duan, Yingjie Zhou, Xiongkuo Min, Qi~Jia, Dongzhan Zhou, Wenlong Zhang, Jiezhang Cao, Xue Yang, Junzhi Yu, Songyang Zhang, Haodong Duan, and Guangtao Zhai.
\newblock Large multimodal models evaluation: A survey.
\newblock \url{https://github.com/aiben-ch/LMM-Evaluation-Survey}, 2025{\natexlab{b}}.
\newblock Project Page: AIBench, available online.

\bibitem[Zhang et~al.(2025{\natexlab{c}})Zhang, Zhao, Fang, Li, Liu, Min, Duan, Chen, and Zhai]{Redundancy}
Zicheng Zhang, Xiangyu Zhao, Xinyu Fang, Chunyi Li, Xiaohong Liu, Xiongkuo Min, Haodong Duan, Kai Chen, and Guangtao Zhai.
\newblock Redundancy principles for mllms benchmarks, 2025{\natexlab{c}}.
\newblock URL \url{https://arxiv.org/abs/2501.13953}.

\bibitem[Zheng et~al.(2023)Zheng, Chiang, Sheng, Zhuang, Wu, Zhuang, Lin, Li, Li, Xing, Zhang, Gonzalez, and Stoica]{llmasajudge}
Lianmin Zheng, Wei-Lin Chiang, Ying Sheng, Siyuan Zhuang, Zhanghao Wu, Yonghao Zhuang, Zi~Lin, Zhuohan Li, Dacheng Li, Eric~P. Xing, Hao Zhang, Joseph~E. Gonzalez, and Ion Stoica.
\newblock Judging llm-as-a-judge with mt-bench and chatbot arena, 2023.
\newblock URL \url{https://arxiv.org/abs/2306.05685}.

\end{thebibliography}
\bibliographystyle{iclr2026_conference}

\newpage
\appendix

\section{Dataset Details}
\label{sec:dataset_details}

This section presents the methodology for constructing the Q-Mirror-Expert and Q-Mirror-Grad datasets. The design emphasizes quality, diversity, and evaluative power to enable rigorous multi-modal assessment.

\paragraph{Phase 1: Candidate Pool Construction}
We begin by sampling candidate TQAs from multiple authoritative scientific benchmarks\citep{rein2024gpqa,eese} to ensure disciplinary breadth and diversity. An LMM automatically filters questions, discarding those without clear visualizable components. Human experts then review the remaining items to verify their scientific accuracy, refine their formulations, and guarantee balanced coverage across 22 scientific disciplines. This process yields a curated pool of 440 high-quality TQAs.

\paragraph{Phase 2: Difficulty-based Partitioning}
The curated pool is then partitioned into two subsets according to difficulty, determined jointly by dataset-provided difficulty labels and expert judgment.

\subparagraph{Q-Mirror-Expert (Expert-level Subset)}
This subset consists of 310 questions (approximately 70\% of the pool). It emphasizes \textbf{novelty, abstraction, and creativity}, including frontier and interdisciplinary concepts, highly abstract problems, and cases where no canonical visual representation exists. It thus evaluates higher-order reasoning and creative inference.

\subparagraph{Q-Mirror-Grad (Graduate-level Subset)}
This subset contains 130 questions (approximately 30\% of the pool). It emphasizes \textbf{structure, established knowledge, and accuracy}, covering mature scientific concepts with canonical visual forms such as molecular structures, circuit diagrams, or biological processes. It thereby measures fidelity in reproducing structured scientific information. The GPQA IDs included in this subset are listed in Table~\ref{tab:gpqa_ids}.

\begin{table}[h]\small
\centering
\caption{GPQA Question IDs Included in Q-Mirror-Grad.}
\label{tab:gpqa_ids}
\begin{tabular}{@{}ccccccccccccccc@{}}
\toprule
\multicolumn{15}{c}{\textbf{GPQA IDs}} \\ \midrule
0 & 1 & 2 & 4 & 7 & 8 & 9 & 12 & 13 & 14 
&15 & 16 & 17 & 18 & 19 \\
20 &22  &23 & 24 & 25  &27 & 28 & 30 & 32 & 33 & 35 & 36 & 37 & 38 & 40 \\
41 & 42 &43 & 44 & 45 & 46 & 47 & 48 & 49 & 50  &52 & 54 & 55 & 56 & 58 \\
60 & 61 & 62 &63 & 64 &66 &67 & 68 & 71 & 72 & 74 & 75 & 76 & 77 & 78 \\
80 & 81 & 82 & 84  &85 & 86 & 87 & 88
 &89 & 90 &91 & 92 & 93 & 94 & 95\\
96 & 97 & 98 & 99 & 100 &101 & 102 & 105 & 107 & 109
&110 & 112 & 113 & 114 & 115 \\
116 & 117 & 118 & 120 & 121 & 123 & 125 & 126 & 127 & 128 &130 & 131 & 132 & 138 & 140\\ 
141 &142 & 144 & 145 & 147 &148 & 149 & 150 & 152 & 153 & 154 & 155 & 157 & 158 & 159\\ 
160 &161 & 162 & 163 & 168 & 169 & 174 & 175 &191 & 194 \\ \bottomrule
\end{tabular}
\end{table}

\begin{figure}[!t]
    \centering
    \includegraphics[width=\linewidth]{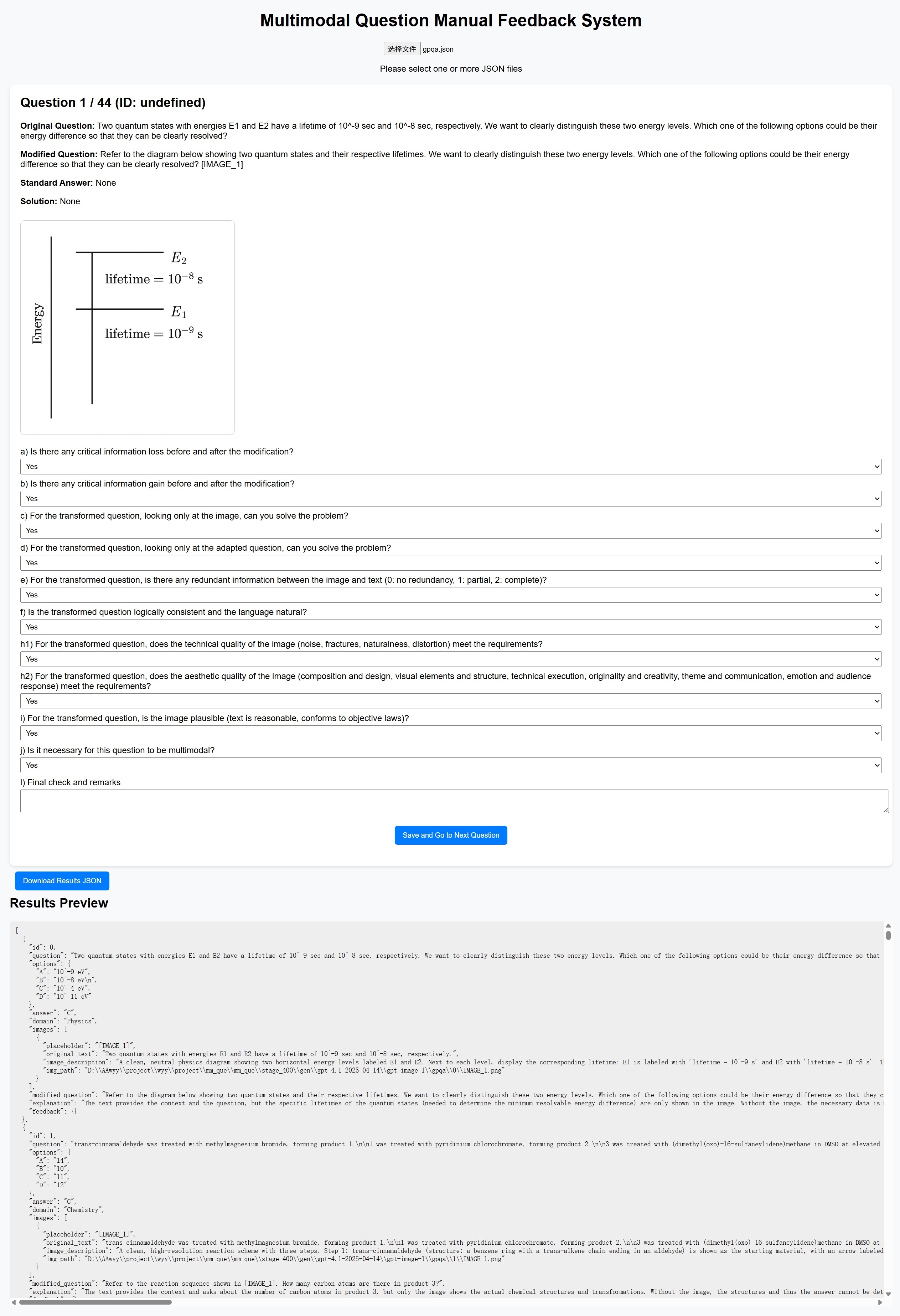}
    \caption{Custom-built web interface for human evaluation. Annotators view the original TQA and the transformed MMQA (text and image), then score the MMQA according to fine-grained metrics. The interface enforces randomized assignment, and automatically exports results in JSON format to ensure independence and reproducibility.}
    \label{fig:biaozhu}
\end{figure}

\section{Human Evaluation Protocol}
\label{sec:human_eval_protocol}

This section details the protocol employed for human expert evaluation, which serves as the gold standard for assessing MMQA quality and benchmarking LMMs as judges.

\paragraph{Annotator Background and Training}
We recruit a panel of 50 annotators, all of whom are Ph.D. students in STEM disciplines with undergraduate training in fields such as physics, chemistry, biology, or mathematics. Each annotator undergoes a standardized 4-hour training session, which includes:
\begin{itemize}[leftmargin=2em]
    \item A comprehensive review of the formal MMQA quality rubric and its underlying principles (Section~\ref{sec:rubric}).
    \item An analysis of high- and low-quality MMQA examples illustrating common pitfalls such as information loss, redundant visuals, and factual inaccuracies.
    \item Practice rounds on 20 sample MMQAs, followed by a group discussion and calibration session to ensure consistent interpretation and application of scoring guidelines.
\end{itemize}
Periodic calibration sessions are further conducted to maintain consistency throughout the evaluation process.

\paragraph{Evaluation Procedure}
Each model-generated MMQA is randomly assigned to two annotators, with a third annotator included in cases of disagreement. Annotators use a custom-built web interface (Figure~5) that presents the original TQA alongside the transformed MMQA (text and image). The interface enforces randomized assignment and prevents annotators from viewing others’ scores, thereby ensuring independence. Annotators score each MMQA according to 11 fine-grained metrics grouped under three principles: Information Consistency (IC), Cross-Modal Integration (CM), and Standalone Quality (QT). For any score below the maximum, annotators also provide brief textual justifications.

\subsection*{Principle 1: Information Consistency (IC)}
This principle measures the semantic fidelity of the transformed MMQA relative to the source TQA.

\emph{Metric 1.1: Information Loss (IL / $p_{-}$)}
\begin{itemize}[leftmargin=2em]
    \item {Definition:} Assesses whether any critical information from the original TQA is missing in the MMQA.
    \item {Scoring (Binary):} 1 = No information loss, 0 = Critical information missing.
\end{itemize}

\emph{Metric 1.2: Information Addition (IA / $p_{+}$)}
\begin{itemize}[leftmargin=2em]
    \item {Definition:} Assesses whether spurious or non-inferable information is introduced.
    \item {Scoring (Binary):} 1 = No spurious information, 0 = Spurious information added.
\end{itemize}

\subsection*{Principle 2: Cross-Modal Integration (CM)}
This principle evaluates whether text and image provide complementary and indispensable information.

\emph{Metric 2.1: Solvability with Image (SI / $p_{s}(I)$)}
\begin{itemize}[leftmargin=2em]
    \item {Definition:} Tests whether the problem can be solved using the image alone.
    \item {Scoring (Binary):} 1 = Not solvable with image alone (desirable), 0 = Solvable with image alone.
\end{itemize}

\emph{Metric 2.2: Solvability with Question (SQ / $p_{s}(T)$)}
\begin{itemize}[leftmargin=2em]
    \item {Definition:} Tests whether the problem can be solved using the text alone.
    \item {Scoring (Binary):} 1 = Not solvable with text alone (desirable), 0 = Solvable with text alone.
\end{itemize}

\emph{Metric 2.3: Redundancy-Synergy (RE / $f_{s}$)}
\begin{itemize}[leftmargin=2em]
    \item {Definition:} Measures the degree of information overlap between text and image.
    \item {Scoring (Ordinal):} 1 = Synergistic, no redundancy, 50 = Partial redundancy, 0 = Complete redundancy.
\end{itemize}

\subsection*{Principle 3: Standalone Quality (QT)}
This principle assesses the intrinsic quality of each modality independently.

\emph{Metric 3.1: Natural Expression (NE / $p_{\text{nat}}$)}
\begin{itemize}[leftmargin=2em]
    \item {Definition:} Evaluates the fluency, coherence, and grammatical quality of the text.
    \item {Scoring (Binary):} 1 = Fluent and correct, 0 = Contains significant linguistic issues.
\end{itemize}

\emph{Metric 3.2: Technical Quality (TQ / $p_{\text{tech}}$)}
\begin{itemize}[leftmargin=2em]
    \item {Definition:} Evaluates whether the image is technically sound and artifact-free.
    \item {Scoring (Binary):} 1 = Technically flawless, 0 = Contains noticeable defects.
\end{itemize}

\emph{Metric 3.3: Aesthetic Quality (AQ / $p_{\text{aes}}$)}
\begin{itemize}[leftmargin=2em]
    \item {Definition:} Evaluates the composition, layout, and visual appeal of the image.
    \item {Scoring (Binary):} 1 = Well-composed and professional, 0 = Poorly composed or unappealing.
\end{itemize}

\emph{Metric 3.4: Semantic Clarity (SC / $p_{\text{sem}}$)}
\begin{itemize}[leftmargin=2em]
    \item {Definition:} Evaluates whether the image is unambiguous, scientifically accurate, and factually correct.
    \item {Scoring (Binary):} 1 = Accurate and unambiguous, 0 = Contains errors or ambiguity.
\end{itemize}

\paragraph{Inter-Annotator Agreement (IAA)}
To quantify the reliability of human judgments, we compute Krippendorff's Alpha ($\alpha$) across all metrics prior to consensus resolution. We achieve a high average agreement of $\alpha = 0.82$, validating the robustness of the protocol and the reliability of human annotations as a gold standard. The evaluation interface (Figure~\ref{fig:biaozhu}) further supports this process by enforcing independent scoring and standardized output, thereby minimizing bias and ensuring reproducibility.

\section{Prompts for Generation}
\label{sec:prompt_generation}

This section provides the full set of prompt templates used for instructing the MMQA generation models. This template is used to guide the multi-modal question modification assistant to generate image-related fields. 

\textbf{\textit{System Prompt:}}\\
\textit{\#System: You are a professional multi-modal question modification assistant. Your task is to generate the image-related fields for the given multiple-choice question without modifying the original question, options, or answer.\\ 
The modified question must meet the following criteria:\\ 
- Both text and image are required to answer the question correctly. Neither the image alone nor the modified text alone can fully solve the question.\\ 
- The modified text must be clear, high-quality, and complete, but it must be understood in conjunction with the image.\\ 
- The image description must be specific, clear, and comprehensive (while strictly less than 1024 characters) to ensure high-quality image generation and clarity of information.\\ 
- The overall information content and difficulty of the question must remain consistent before and after modification.\\ 
- The visual information should be removed from the modified text, and instead, it should be represented in the image.\\ 
  - For example, if the question previously described an energy difference between two states, the visual representation of the energy difference should be shown in the image, while the text should simply provide the context or question that requires the image for full understanding.\\ 
- The image description should be as detailed as possible (while remaining under 1024 characters), providing specific instructions for creating the image. The description should cover all the visual aspects required to convey the context of the question, ensuring a high-quality visual representation.\\
- The image style and composition must be neutral and objective. The visual style must be appropriate for the question's subject area (e.g., a clean diagram for science, an antique-style illustration for history). Critically, the image's composition, color, and layout must not use artistic techniques (such as special highlights, pointing arrows, or making elements related to one option more prominent) to unintentionally hint at or guide towards the correct answer, to ensure the question's fairness.\\
- The text and visual information between the modified\_question and image\_description should not overlap. The image should visually represent what the text cannot explain or clarify, avoiding any duplication of the same information in both the text and the image.}

Please return the following JSON format:
\{\\
``images": [\\
    \{\\
``placeholder": ``[IMAGE\_1]",\\
   ``original\_text": ``The original text that should be replaced",\\
   ``image\_description": ``A detailed description (must be less than 1024 characters) of the image to be generated"
    \}\\
],\\
``modified\_question": ``The modified question text, containing a placeholder [IMAGE\_1] for the image position. Ensure that the visual content is removed from the text and only context remains.",\\
``explanation": ``A brief explanation of why both the image and text are needed."\\
\}

\section{Prompts for LMM Quality Judge}
\label{sec:prompt_eval}

This section provides the full set of prompt templates used for instructing the LMM quality judges. The prompt to instruct an LMM to act as a quality judge for an MMQA is designed to elicit structured scores and justifications based on our rubric. This template is used to guide the professional exam quality reviewer to evaluate multi-modal questions across different stages.

\textbf{Stage 1: Evaluate information loss}

\textit{\#System: You are a professional exam quality reviewer. Your task is to evaluate whether there is any information loss between the original and transformed question. Respond strictly as instructed.\\
Evaluation schema:
{
  a) Is there any critical information loss before and after the modification?
}\\
Original question:
``{item['question']}"\\
Modified question:
``{item['modified\_question']}"\\
Added image:
``{item['image\_url']}"
}

Please answer strictly in the following format:\\
\{
a) [yes/no]\\
l) [your comments explaining the reasoning behind your answer to a)]
\}

\textbf{Stage 2: Evaluate information addition}

\textit{\#System: You are a professional exam quality reviewer. Your task is to evaluate whether there is any information addition between the original and transformed question. Respond strictly as instructed.\\
Evaluation schema:
{
  b) Is there any critical information gain before and after the modification?
}\\
Modified question:
``{item['modified\_question']}"\\
Added image:
``{item['image\_url']}"
}

Please answer strictly in the following format:\\
\{b) [yes/no]\\
l) [your comments explaining the reasoning behind your answer to b)]
\}

\textbf{Stage 3: Evaluate solvability with image only}

\textit{\#System: You are a professional exam quality reviewer. Your task is to evaluate whether the transformed question can be solved using only the image and options. Respond strictly as instructed.\\
Evaluation schema:
{
  c) For the transformed question, looking only at the image and options, can you solve the problem?
}\\
Added image:
``{item[`image\_url']}"\\
(if have) Options:
``{item[`question']}"
}

Please answer strictly in the following format:\\
\{c) [yes/no]\\
l) [your comments explaining the reasoning behind your answer to c)]
\}

\textbf{Stage 4: Evaluate solvability with text only}

\textit{\#System: You are a professional exam quality reviewer. Your task is to evaluate whether the transformed question can be solved using only the modified question text and options. Respond strictly as instructed.\\
Evaluation schema:
{
   d) For the transformed question, looking only at the transformed question and options, can you solve the problem?
}\\
Modified question:
``{item[`modified\_question']}"\\
(if have) Options:
``{item[`question']}"
}

Please answer strictly in the following format:\\
\{d) [yes/no]\\
l) [your comments explaining the reasoning behind your answer to d)]
\}

\textbf{Stage 5: Evaluate information redundancy}

\textit{\#System: You are a professional exam quality reviewer. Your task is to evaluate the level of redundant information between the image and text in the transformed question. Respond strictly as instructed.\\
Evaluation schema:
{
  e) For the transformed question, is there any redundant information between the image and text (0: no redundancy, 1: partial, 2: complete)?
}\\
Original question:
``{item[`question']}"\\
Modified question:
``{item[`modified\_question']}"\\
Added image:
``{item[`image\_url']}"\\
}

Please answer strictly in the following format:\\
\{e) [0/1/2] (0: no redundancy, 1: partial redundancy, 2: complete redundancy)\\
l) [your comments explaining the reasoning behind your answer to e)]
\}

\textbf{Stage 6: Evaluate logical consistency and language naturalness}

\textit{\#System: You are a professional exam quality reviewer. Your task is to evaluate whether the transformed question is logically consistent and has natural language. Respond strictly as instructed.\\
Evaluation schema:
{
   f) Is the transformed question logically consistent and the language natural?
}\\
Modified question:
``{item[`modified\_question']}"\\
}

Please answer strictly in the following format:\\
\{
f) [yes/no]\\
l) [your comments explaining the reasoning behind your answer to f)]
\}

\textbf{Stage 7: Evaluate image technical quality}

\textit{\#System: You are a professional exam quality reviewer. Your task is to evaluate whether the technical quality of the image meets the requirements. Respond strictly as instructed.\\
Evaluation schema:
{
  h1) For the transformed question, does the technical quality of the image (noise, fractures, naturalness, distortion) meet the requirements?
}\\
Added image:
``{item[`image\_url']}"\\
}

Please answer strictly in the following format:\\
\{
h1) [yes/no]\\
l) [your comments explaining the reasoning behind your answer to h1)]
\}

\textbf{Stage 8: Evaluate image aesthetic quality}

\textit{\#System: You are a professional exam quality reviewer. Your task is to evaluate whether the aesthetic quality of the image meets the requirements. Respond strictly as instructed.\\
Evaluation schema:
{
  h2) For the transformed question, does the aesthetic quality of the image (composition and design, visual elements and structure, technical execution, originality and creativity, theme and communication, emotion and audience response) meet the requirements?
}\\
Added image:
``{item[`image\_url']}"\\
}

Please answer strictly in the following format:\\
\{
h2) [yes/no]\\
l) [your comments explaining the reasoning behind your answer to h2)]
\}

\textbf{Stage 9: Evaluate semantical clarity}

\textit{\#System: You are a professional exam quality reviewer. Your task is to evaluate whether the image is clear and conforms to objective laws. Respond strictly as instructed.\\
Evaluation schema:
{
  i) For the transformed question, is the image plausible (text is reasonable, conforms to objective laws)?
}\\
Added image:
``{item[`image\_url']}"\\
}

Please answer strictly in the following format:\\
\{
i) [yes/no]\\
l) [your comments explaining the reasoning behind your answer to i)]
\}

\section{Experiments}
\label{sec:app_experiments}

\subsection{Experiment Analysis.}
\begin{figure}[h]
    \centering
    \includegraphics[width=\linewidth]{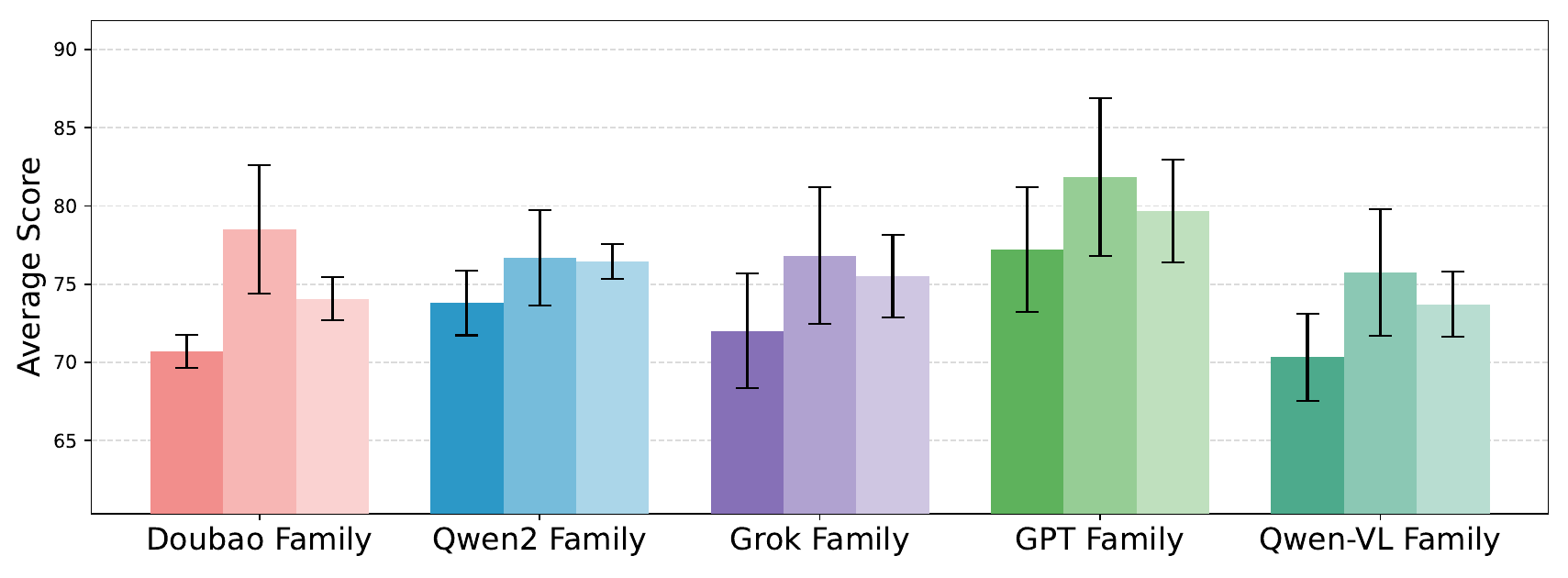}
    \caption{Robustness performance on MMQA generation Benchmark. Each bar shows the mean AVG score with variance, and the three bars from left to right indicate results on Q-Mirror-(Expert+Grad), Q-Mirror-Expert, and Q-Mirror-Grad.}
    \label{fig:score}
\end{figure}

\noindent\textbf{Findings for MMQA Generation.} 
Examining per-metric behavior (see \S\ref{sec:principle}), we observe consistently strong modal interdependence, reflected in high SI and SQ scores across model families. This confirms that our TQA-to-MMQA transformation enforces genuine cross-modal reasoning rather than allowing text-only shortcuts. By contrast, factual fidelity in the image space (SC) remains the principal limitation on the Expert subset, where mislabeled components and subtle schematic inaccuracies frequently occur. These findings indicate that one-shot generation is suboptimal for expert-level MMQA: models align modalities well but fail to ground fine-grained visual details reliably. This trend is further illustrated in Figure~\ref{fig:score}, which compares performance across model families under varying weight assignments for the three core evaluation dimensions (IC, CM, QT). The results show that while models remain stable under changes emphasizing modal interdependence (CM), their scores drop significantly when factual accuracy (IC/SC) receives higher weight, especially on Q-Mirror-Expert. This confirms that factual grounding in the visual modality is the dominant bottleneck in current one-shot generation approaches.

\noindent\textbf{Findings for MMQA Evaluation.} 
Analysis of model-as-judge performance reveals three consistent trends. First, judge accuracy is stratified: Grok-4 achieves the highest agreement with human annotations (62.8\%), while weaker models remain below 55\%. Second, evaluation difficulty varies by abstraction level. All judges align more closely with humans on Q-Mirror-Grad than on Q-Mirror-Expert. For example, Claude-Sonnet-4 improves from 58.98\% on Expert to 71.00\% on Grad, reflecting that graduate-level tasks admit more objective criteria, whereas expert-level tasks involve inherently subjective reasoning. Third, correlations among major quality dimensions (IC, CM, QT) are low, validating the non-redundancy of the rubric design and affirming the effectiveness of the underlying evaluation principles. These findings underscore the risk of relying on a single judge and provide motivation for adopting an ensemble of high-performing models to obtain more stable and unbiased evaluation.

\noindent\textbf{Q-Mirror Agent Performance Improvement.} 
The Q-Mirror agent yields substantial improvements in MMQA quality. Across all 440 scientific questions, the overall score rises by $\bar{\Delta}=6.32$ points (from 78.90 to 85.22). More importantly, the pass rate increases from 72\% to 95\%. This demonstrates that the agent effectively identifies and corrects flaws in nearly one-third of initially suboptimal generations. In side-by-side comparisons, human evaluators prefer agent-refined outputs 78\% of the time. Notably, gains are most pronounced on the Expert subset, confirming that iterative refinement is particularly valuable for complex, less standardized scientific problems.

\begin{figure}[t]
    \centering
    \includegraphics[width=\linewidth]{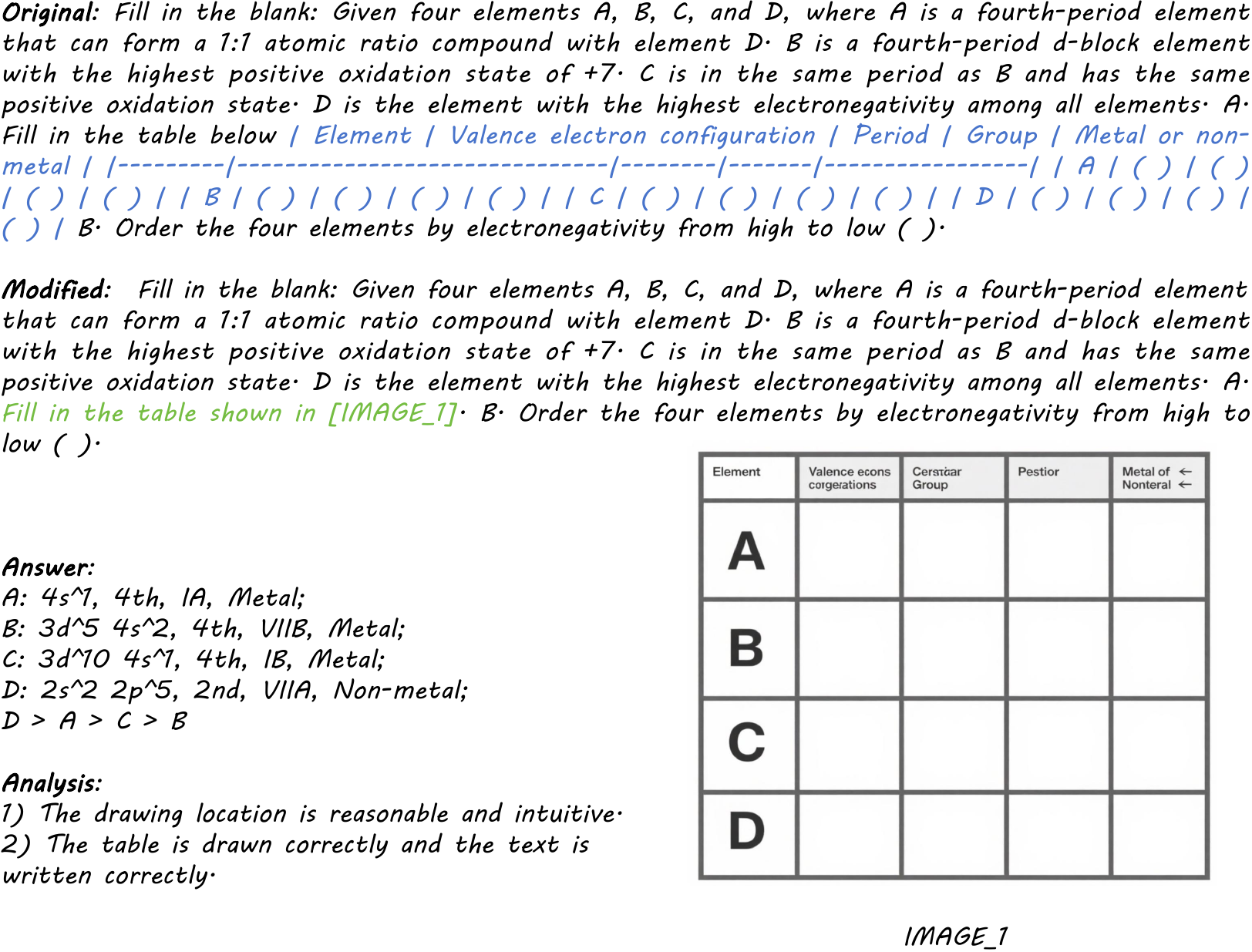}
    \caption{A success case where the Q-Mirror agent correctly transforms a text-heavy chemistry problem into a structured visual table. The visual representation is accurate, the textual formulation remains fluent, and the modalities exhibit strong synergy, collectively preserving information consistency.}
    \label{fig:success1}
\end{figure}

\begin{figure}[t]
    \centering
    \includegraphics[width=\linewidth]{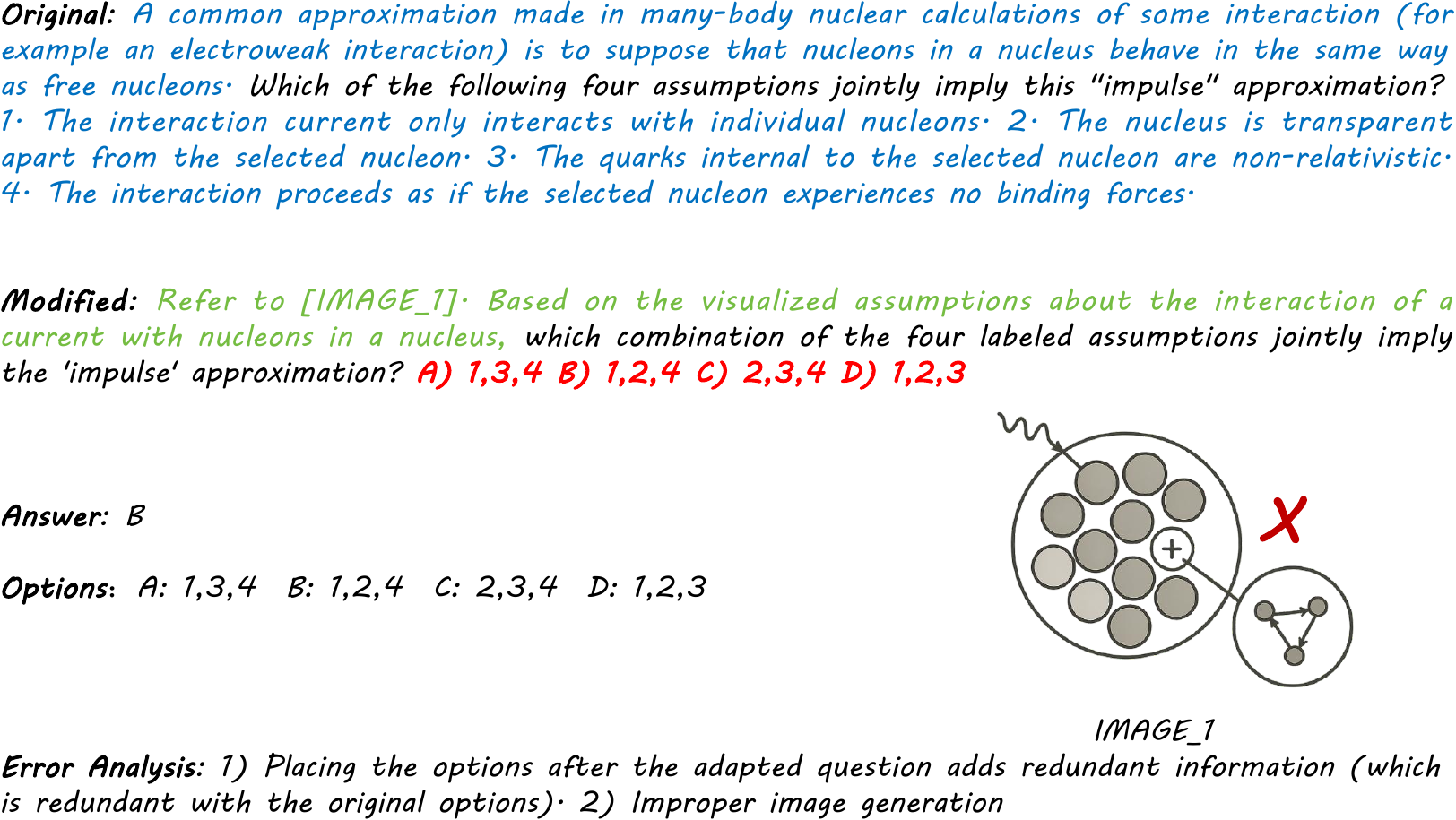}
    \caption{A failure case involving abstract nuclear physics assumptions. The agent attempts to visualize four principles (e.g., `transparent nucleus'), but produces a generic, ambiguous diagram. The image lacks semantical clarity, offers little complementary information, and results in redundancy with the text.}
    \label{fig:error1}
\end{figure}

\begin{figure}[t]
    \centering
    \includegraphics[width=\linewidth]{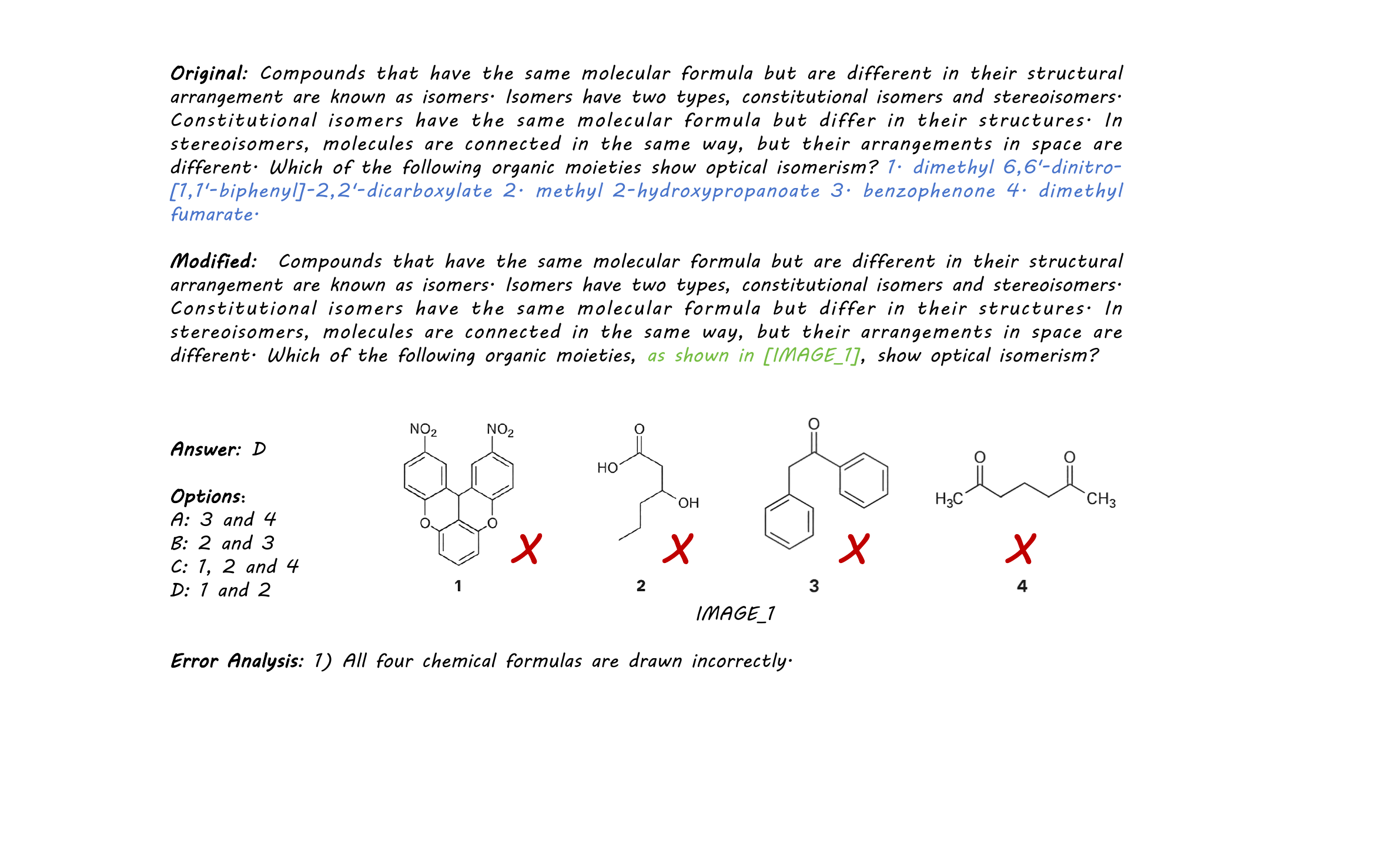}
    \caption{A failure case in organic chemistry, where the agent is required to draw molecular structures for four compounds. All structures are chemically incorrect, resulting in factual inconsistency and a breakdown of technical quality.}
    \label{fig:error2}
\end{figure}

\begin{figure}[!t]
    \centering
    \includegraphics[width=\linewidth]{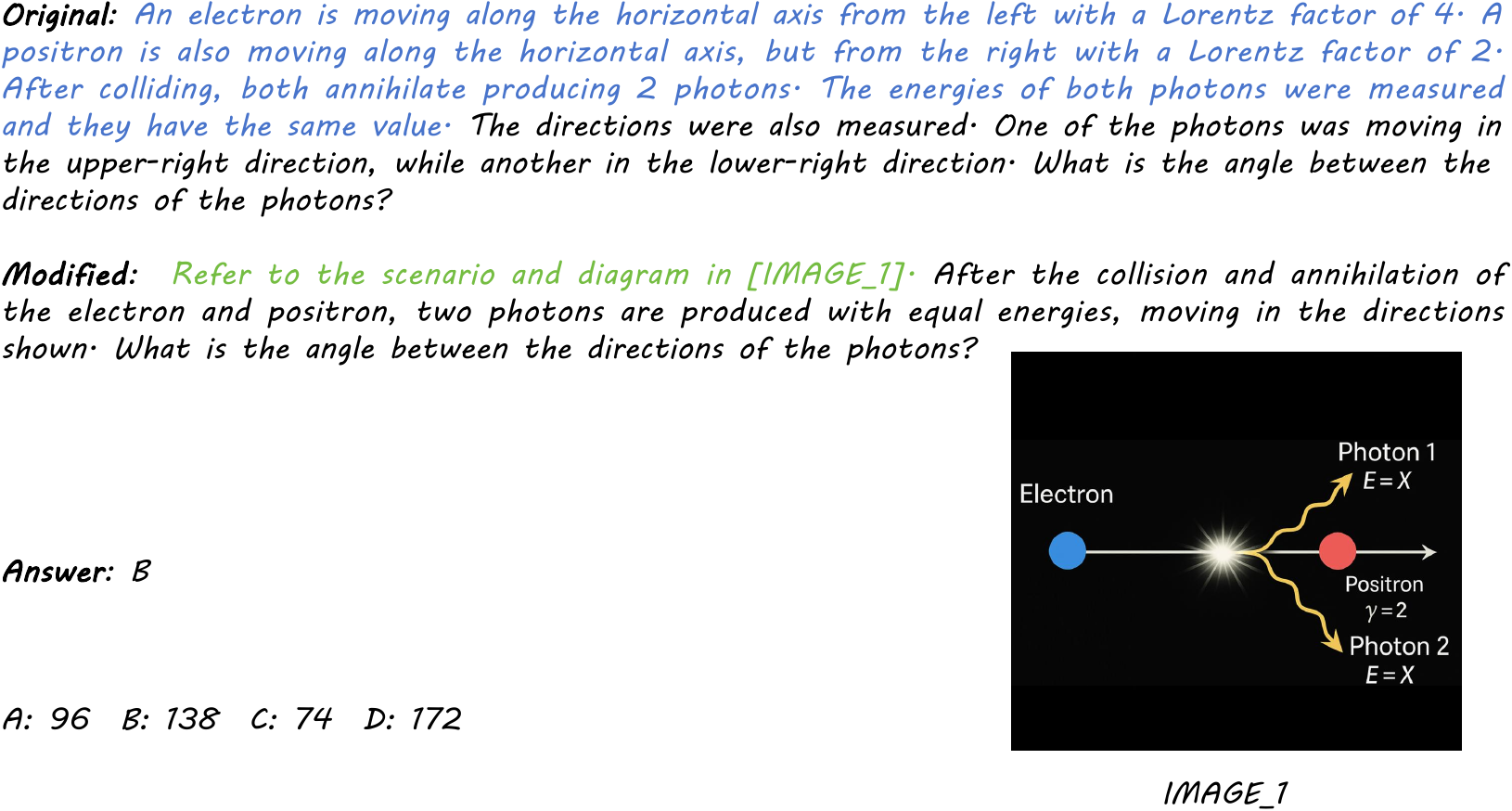}
    \caption{A success case highlights the transformation of a scientific TQA into its multi-modal counterpart (MMQA). The upper part shows the original TQA, while the lower part displays the O-Mirror generated MMQA, where key textual cues have been converted into a visual representation. The example illustrates how O-Mirror preserves semantic fidelity, introduces a figure to enforce multi-modal dependence, and maintains overall quality and clarity.}
    \label{fig:agent_case_1}
\end{figure}

\begin{figure}[!t]
    \centering
    \includegraphics[width=\linewidth]{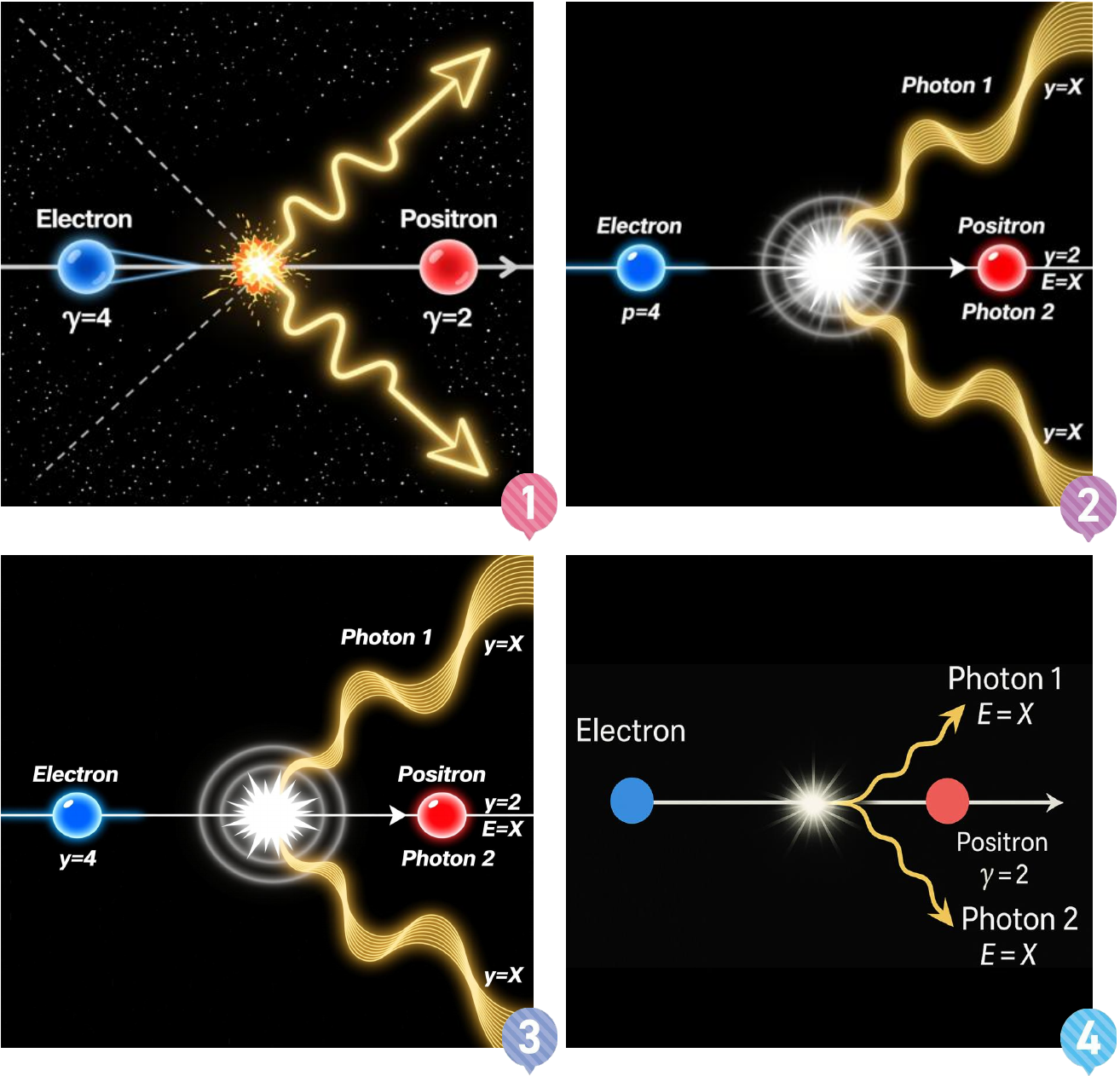}
    \caption{ Step-by-step illustration of the Q-Mirror Agent’s iterative generation process for the same scientific case shown in Figure~\ref{fig:agent_case_1}. 
        (1) \textbf{Initial modal conversion:} The original text-only QA pair (TQA) is converted into a visualizable event description, where an electron and a positron collide and produce two photons. 
        (2) \textbf{Raw image generation:} A first draft of the multi-modal QA pair (MMQA) is produced, integrating particle labels and energy/momentum annotations, though with incomplete clarity. 
        (3) \textbf{Iterative refinement:} Based on evaluator feedback, the image layout and labeling are corrected to enforce physical symmetry and information consistency. 
        (4) \textbf{Final polished output:} A simplified and scientifically faithful diagram is generated, achieving clarity, semantic fidelity, and aesthetic quality.}
    \label{fig:agent_case_2}
\end{figure}

\newpage 

\subsection{Case Study Analysis of the Q-Mirror Agent}
\label{sec:qmirror_case_studies}

To further assess the practical performance of the Q-Mirror agent, we present representative case studies spanning both successful and failed generations. These examples provide a nuanced view of the agent’s capabilities and limitations, evaluated through our quality rubric (IC, CM, and QT). In addition to static outcomes, we also examine the iterative dynamics of Q-Mirror’s closed-loop pipeline, thereby offering a comprehensive understanding of its operational behavior.

\subsubsection{Success Case: Structuring Implicit Information}

\noindent\textbf{Context.} This case involves a chemistry problem that requires filling in a table, where the table’s structure is described only in text (Figure~\ref{fig:success1}).

\begin{itemize}[leftmargin=2em]
    \item \textbf{Transformation:} The Q-Mirror agent parses the textual description of rows and columns and successfully converts the implicit structure into an explicit, well-formatted table in the generated image, while retaining the original textual problem statement.
    \item \textbf{Evaluation:} 
    \textbf{IC:} All scientific information is faithfully preserved.  
    \textbf{CM:} Both modalities contribute indispensably: the text conveys scientific context, while the image encodes structural information, yielding strong synergy.  
    \textbf{QT:} The table is clear, unambiguous, and adheres to conventional scientific presentation standards.  
\end{itemize}

This case exemplifies a prototypical success, in which Q-Mirror effectively externalizes structural representations into the visual modality without sacrificing fidelity.

\subsubsection{Success Case 2}

To complement the Q-Mirror analysis, we provide a detailed case study of the \textbf{O-Mirror agent} (Figure~\ref{fig:agent_case_1}). Unlike aggregate benchmarks, this example traces the full pipeline of a single transformation from a text-only QA pair (TQA) to a multi-modal QA pair (MMQA).

\paragraph{Before (TQA).} 
The question is entirely text-based, requiring readers to infer visual or structural elements implicitly.

\paragraph{After (MMQA).} 
O-Mirror automatically extracts latent visual cues, generates a scientifically aligned illustration, and integrates it with the reformulated question.

This case highlights three critical design principles of O-Mirror:
\begin{enumerate}
    \item \textbf{Preservation of Meaning}: No essential information is lost or distorted in the transformation.
    \item \textbf{Cross-Modal Integration}: The multimodal question requires reasoning across both text and image, preventing shortcut solutions.
    \item \textbf{Improved Clarity}: The visual representation enhances comprehension while remaining scientifically plausible.
\end{enumerate}

Through this case study, we demonstrate the interpretability and effectiveness of O-Mirror’s pipeline, offering concrete evidence that the framework can faithfully adapt scientific questions into high-quality multi-modal formats.

\subsubsection{Failure Mode 1: Ambiguous Visualization of Abstract Concepts}

\noindent\textbf{Context.} This case is drawn from a nuclear physics problem concerning the `impulse approximation', where the agent must visualize four abstract assumptions (Figure~\ref{fig:error1}).

\begin{itemize}[leftmargin=2em]
    \item \textbf{Transformation:} The agent produces a generic schematic that attempts to capture multiple abstract principles (e.g., `the nucleus is transparent' and `no binding forces'). However, the resulting image fails to distinguish among the assumptions in a meaningful way.
    \item \textbf{Evaluation:} 
    \textbf{IC:} The textual content is preserved, but the visual contributes little additional meaning.  
    \textbf{CM:} The lack of discriminative visual elements leads to weak cross-modal integration.  
    \textbf{QT:} The image scores poorly on \textit{Semantic Clarity} ($p_{\text{sem}}$), as the intended concepts are not effectively conveyed.  
\end{itemize}

This highlights Q-Mirror’s difficulty in translating abstract conceptual descriptions into precise, scientifically interpretable visualizations.

\subsubsection{Failure Mode 2: Factual Inaccuracy in Technical Diagrams}

\noindent\textbf{Context.} This case concerns an organic chemistry problem on optical isomerism, where the agent is tasked with generating molecular structures for four compounds (Figure~\ref{fig:error2}).

\begin{itemize}[leftmargin=2em]
    \item \textbf{Transformation:} The agent produces four diagrams intended as molecular structures, all of which are chemically incorrect.
    \item \textbf{Evaluation:} 
    \textbf{IC:} The original entities are not preserved, resulting in a complete violation of information consistency.  
    \textbf{CM:} As the diagrams are invalid, the image does not contribute to reasoning; the text alone suffices.  
    \textbf{QT:} The diagrams score zero for \textit{Technical Quality} ($p_{\text{tech}}$) due to factual inaccuracies.  
\end{itemize}

This case underscores that factual correctness is non-negotiable in scientific domains and highlights the need for domain-aware verification mechanisms to safeguard against technically invalid outputs.

\subsubsection{Iterative Generation Dynamics}

While the above cases highlight static outcomes, Figure~\ref{fig:agent_case_2} illustrates how the Q-Mirror agent progressively refines a text-only QA pair (TQA) into a high-quality multi-modal QA pair (MMQA).

\begin{itemize}[leftmargin=2em]
    \item \textbf{Step (1): Initial conversion} identifies the electron:positron annihilation event but provides only a conceptual representation with limited labeling. 
    \item \textbf{Step (2): Raw image generation} introduces particle annotations and energy values, though the layout remains cluttered and lacks physical symmetry. 
    \item \textbf{Step (3): Iterative refinement} incorporates evaluator feedback to correct photon emission directions and reduce redundant text, thereby improving information consistency. 
    \item \textbf{Step (4): Final output} converges to a polished diagram that balances semantic fidelity, cross-modal integration, and visual clarity. 
\end{itemize}

This example demonstrates the internal dynamics of Q-Mirror’s closed-loop generation pipeline, where planner, evaluator, and controller components interact to progressively enhance fidelity, alignment, and presentation quality.

\end{document}